\documentclass[letterpaper]{article} % DO NOT CHANGE THIS
\usepackage{aaai24}  % DO NOT CHANGE THIS
\usepackage{times}  % DO NOT CHANGE THIS
\usepackage{helvet}  % DO NOT CHANGE THIS
\usepackage{courier}  % DO NOT CHANGE THIS
\usepackage[hyphens]{url}  % DO NOT CHANGE THIS
\usepackage{graphicx} % DO NOT CHANGE THIS
\usepackage{amsmath}
\usepackage{natbib}  % DO NOT CHANGE THIS AND DO NOT ADD ANY OPTIONS TO IT
\usepackage{caption} % DO NOT CHANGE THIS AND DO NOT ADD ANY OPTIONS TO IT
\usepackage{algorithm}
\usepackage{amsfonts}
\usepackage{amsmath}
\usepackage{algpseudocode}
\usepackage{svg}
\usepackage{xspace}
\usepackage{multirow}
\usepackage{booktabs}
\usepackage{colortbl}
\usepackage{algpseudocode}
\usepackage{color, colortbl}

\usepackage[skins]{tcolorbox}
\usepackage{subfig}

\usepackage[]{tcolorbox}        

\tcbset{
    colback = white,
    before skip = 0.2cm,    
    after skip = 0.5cm      
}                           % setting global options for tcolorbox

\newtcolorbox{boxA}{
    boxrule = 0.2pt,
    colframe = black,
    colback=yellow!10!white% frame color
}

\newcommand{\OUR}{RSGG-CE\xspace}
\newcommand{\LONGOUR}{\textbf{R}obust \textbf{S}tochastic \textbf{G}raph \textbf{G}enerator for \textbf{C}ounterfactual   \textbf{E}xplanations\xspace}

\usepackage{newfloat}
\usepackage{listings}
\DeclareCaptionStyle{ruled}{labelfont=normalfont,labelsep=colon,strut=off} % DO NOT CHANGE THIS
\floatstyle{ruled}
\newfloat{listing}{tb}{lst}{}
\floatname{listing}{Listing}
\title{Robust Stochastic Graph Generator for Counterfactual Explanations}
\author{Mario Alfonso Prado-Romero\equalcontrib\textsuperscript{\rm 1}, Bardh Prenkaj\equalcontrib\textsuperscript{\rm 2}, Giovanni Stilo\textsuperscript{\rm 3}}
\affiliations{
    \textsuperscript{\rm 1} Gran Sasso Science Institute
    \textsuperscript{\rm 2} Sapienza University of Rome
    \textsuperscript{\rm 3} University of L'Aquila
    \\
    marioalfonso.prado@gssi.it, prenkaj@di.uniroma1.it, giovanni.stilo@univaq.it\\
    \textcolor{blue}{\url{https://github.com/MarioTheOne/GRETEL}}
}

\usepackage{bibentry}
\begin{document}

\maketitle

\begin{abstract}
Counterfactual Explanation (CE) techniques have garnered attention as a means to provide insights to the users engaging with AI systems. While extensively researched in domains such as medical imaging and autonomous vehicles, Graph Counterfactual Explanation (GCE) methods have been comparatively under-explored. GCEs generate a new graph similar to the original one, with a different outcome grounded on the underlying predictive model. Among these GCE techniques, those rooted in generative mechanisms have received relatively limited investigation despite demonstrating impressive accomplishments in other domains, such as artistic styles and natural language modelling. The preference for generative explainers stems from their capacity to generate counterfactual instances during inference, leveraging autonomously acquired perturbations of the input graph. Motivated by the rationales above, our study introduces \OUR, a novel \LONGOUR able to produce counterfactual examples from the learned latent space considering a partially ordered generation sequence. Furthermore, we undertake quantitative and qualitative analyses to compare \OUR's performance against SoA generative explainers, highlighting its increased ability to engendering plausible counterfactual candidates.
\end{abstract}

\section{Introduction}\label{sec:introduction}
Explainability is crucial in sensitive domains to enable users and service providers to make informed and reliable decisions \cite{guidotti2018survey}. However, deep neural networks, commonly used for generating predictions, often suffer from a lack of interpretability, widely referred to as the \textit{black-box} problem \cite{petch2021opening}, hindering their wide adoption in domains such as healthcare and finance. On the other end of the spectrum of explainability, we find inherently interpretable {white-box} prediction models \cite{8882211}, which are preferred for decision-making purposes \cite{verenich2019predicting}. Alas, black-box models demonstrate superior performance and generalisation capabilities when dealing with high-dimensional data \cite{aragona2021coronna,ding2019effective,feng2019understanding,huang2020skipgnn,madeddu2020feature,prenkaj2021hidden,prenkaj2020reproducibility,prenkaj2023self,verma2022temporal,wang2017deep}.

Recently, deep learning (relying on GNNs \cite{scarselli2008graph}) has been beneficial in solving graph-based prediction tasks, such as community detection \cite{10.1145/3534678.3539370}, link prediction \cite{10.1145/3554981}, and session-based recommendations \cite{wu2019session,DBLP:conf/ijcnn/XuXW21}. Despite their remarkable performance, GNNs are black boxes, making them unsuitable for high-impact and high-risk scenarios. The literature has proposed several post-hoc explainability methods to understand \textit{what is happening under the hood} of the prediction models. Specifically, counterfactual explainability is useful to understand how modifications in the input lead to different outcomes. Similarly, a recent field in Graph Counterfactual Explainability (GCE) has emerged \cite{prado2022survey}.

We provide the reader with an example that helps clarify a counterfactual example in graphs. Suppose we have a social network where a specific user $U$ posts an illicit advertisement, thus violating the Terms of Service (ToS). A counterfactual explanation of $U$'s account suspension would be {\textit{if the user had refrained from writing the post about selling illegal goods, her account would not have been banned}}.

Generally, GCE methods can be search, heuristic, and learning-based approaches \cite{prado2022survey}. Search-based approaches find counterfactual examples within the data distribution. Heuristic-based approaches perturb the original graph $G$ into $G'$ such that, for a certain prediction model $\Phi$, namely oracle, $\Phi(G) \neq \Phi(G')$ without accessing the dataset $\mathcal{G}$. In other words, $G'$ can be outside the data distribution of $\mathcal{G}'$. Heuristic-based approaches suffer the need to define the perturbation heuristic (e.g., rules), which might come after careful examination of the data and involve domain expertise to express how the input graph should be perturbed faithfully. For instance, producing valid counterfactuals for molecules requires knowledge about atom valences and chemical bonds. Contrarily, learning-based approaches learn the generative ``heuristic'' based on the data. This kind of explainer is trained on samples and thus can be used to produce counterfactual instances at inference time.

In this work, we propose \OUR, a \LONGOUR, to produce counterfactual examples from the learned latent space considering a partially ordered generation sequence. \OUR is not confined to the data distribution (vs. search approaches) and does not rely on a learned mask to apply to the input to produce counterfactuals (vs. other learning approaches). However, it learns 
the perturbation of the input autonomously (vs. heuristic approaches), relying on a partial order generation strategy.
Moreover, it does not need access extensively to the oracle $\Phi$ since its latent space can be sampled to generate multiple candidate counterfactuals for a particular input graph. 
The following discusses the related works in Sec. \ref{sec:related_work}. In Sec. \ref{sec:method}, we present the method and the needed preliminary knowledge (see Sec. \ref{sec:prelimniaries}). Finally, in Sec. \ref{sec:experiments}, we conduct the performance analysis, four ablation studies, and a qualitative anecdotal inspection.

\section{Related Work}\label{sec:related_work}

The literature distinguishes between inherently explainable and black-box methods \cite{guidotti2018survey}. Black-box methods can be further categorised into factual and counterfactual explanation methods. Here, we concentrate on counterfactual methods as categorised in \cite{prado2022survey} and exploit the same notation used in that survey.

While many works provide counterfactual explanations for images/text (to point out some \cite{vermeire2022explainable,xu2023counterfactual,Zemni_2023_CVPR}), only a few focus on graph classification problems \cite{abrate2021counterfactual,liu2021multi,ma2022clear,nguyen2022explaining,numeroso2021meg,tancf2,wellawatte2022model}. According to \cite{prado2022survey}, GCE works are categorised into search (heuristic) and learning-based approaches. We are aware that a new branch of global (model-level) counterfactual explanations is being developed (see \cite{huang2023global}). Here, we treat only instance-level and learning-based explainers.

\paragraph{Learning-based approaches} The methods belonging to this category share a three-step pipeline: \textit{1)} generating masks that indicate the relevant features given a specific input graph $G$; \textit{2)} combining the mask with $G$ to derive a new graph $G'$; \textit{3)} feeding $G'$ to the prediction model (oracle) $\Phi$ and updating the mask based on the outcome $\Phi(G')$.
Learning-based strategies can be divided into perturbation matrix \cite{tancf2}, reinforcement learning \cite{nguyen2022explaining,numeroso2021meg,wellawatte2022model}, and generative approaches \cite{ma2022clear}. After training, the learned latent space of generative approaches can be exploited as a sampling basis to engender plausible counterfactuals. (w.l.o.g.), generative methods learn a latent space that embeds original and non-existing estimated edges' edge probabilities (e.g., see \cite{ma2022clear}). This way, one can employ sampling techniques to produce counterfactual candidates w.r.t. the input instance. In the following, we report the most recent and effective SoA methods.

MEG \cite{numeroso2021meg} and MACCS \cite{wellawatte2022model} employ multi-objective reinforcement learning (RL) models (retrained for each input instance) to generate molecule counterfactuals. Their domain-specificity limits their applicability and makes them difficult to port on other domains. The reward function incorporates a task-specific regularisation term that influences the choice of the next action to perturb the input. MACDA \cite{nguyen2022explaining} uses RL to produce counterfactuals for the drug-target affinity problem.

CF$^2$ \cite{tancf2} balances factual and counterfactual reasoning to generate explanations. Like other factual-based approaches, it identifies a subgraph in the input, then presents the remainder as a counterfactual candidate by removing this subgraph \cite{bajaj2021robust}. Notably, CF$^2$ favours smaller explanations for simplicity.

CLEAR \cite{ma2022clear} uses a variational autoencoder (VAE) to encode the graphs into its latent representation $Z$. The decoder generates counterfactuals based on $Z$, conditioned on the explainee class $c \neq \Phi(G)$. Generated counterfactuals are complete graphs with stochastic edge weights. To ensure validity, the authors employ a sampling process. However, decoding introduces node order differences between $G$ and $G'$. Thus, a graph matching procedure (NP-hard \cite{livi2013graph}) between the two is necessary.

While unrelated to graphs, \cite{nemirovsky22} rely on a GAN. They produce counterfactual candidates by training the generator to elucidate a user-defined class. \cite{prado2023revisiting} adapt this into G-CounteRGAN by treating the adjacency matrix as black-and-white images and employing 2D image convolutions.

\paragraph{What is our contribution to the literature?} We design a novel approach to generate counterfactuals by leveraging the latent space of the generator network to reconstruct the input's topology. The discriminator guides this process, which compels the generator to learn the production of counterfactuals aligned with the opposite class.

Firstly, we tackle the limitations of factual-based methods that remove subgraph components to craft counterfactual candidates, a strategy found in \cite{bajaj2021robust,tancf2}. However, this falters when dual classes clash (e.g., acyclic vs cyclic graphs). In such cases, shifting from cyclic to acyclic mandates edge removal, while acyclic to cyclic requires edge addition\footnote{Creating a cyclic graph from acyclic needs an added edge for a loop.}. Since our method combines the generated residual weighted edges with the original edges, we empower both edge addition and removal operations.

Unlike images, graphs lack node ordering, rendering standard 2D convolutions inadequate due to the significance of node adjacency. We integrate Graph Convolution Networks (GCNs) to address this, naturally capturing node neighbourhoods via message-passing mechanisms  \cite{feng2022messagepassing}.

Our approach is zero-shot counterfactual generation. Prevalent techniques, especially those rooted in Reinforcement Learning (RL) \cite{nguyen2022explaining,numeroso2021meg,wellawatte2022model}, require recalibration at inference time to generate counterfactuals for previously unseen graphs.
Differently, we learn a latent graph representation enabling stochastic estimations of the graph's topology, allowing us to reconstruct and generate counterfactuals without retraining.

Lastly, our model introduces an innovative strategy - i.e., partial-order sampling - using estimated edge probabilities acquired from the generator network. To the best of our knowledge, this is the first work that proposes a partial-order \cite{dan2008mathematical} sampling approach on estimated edges. This aids in identifying sets of edges that should be sampled first for effective counterfactual generation (see Sec. \ref{sec:method} and Algorithm \ref{algo:sampling}).

\section{Preliminaries}\label{sec:prelimniaries}

This section briefly overviews the fundamental concepts and techniques relevant to our study on robust counterfactual explanations. We introduce the concepts of graphs, adjacency matrices, and graph counterfactuals.

\paragraph{Graphs and Adjacency Matrix} A graph, denoted as $G = \left( X,A \right)$, is a mathematical structure consisting of node features $X \in \mathbb{R}^{n \times d}$ and an adjacency matrix $A \in \mathbb{R}^{n\times n}$ which represents the connectivity between nodes. For an undirected weighted graph, the adjacency matrix $A$ is symmetric, and its elements are defined as
\begin{equation}
A \left[ v_i , v_j \right] = 
    \begin{cases}
        w \left( v_i , v_j \right) & \text{if } \left( v_i , v_j \right) \text{ is an edge in } G\\
        0 & \text{otherwise }
    \end{cases}
\end{equation}
where $w \left( v_i , v_j \right) \in \mathbb{R}$ is the weight vector of the edge incident to the nodes $v_i$ and $v_j$. For directed graphs, the adjacency matrix may exhibit asymmetry, thereby indicating the directionality of the edges. We focus on undirected graphs and denote the graph dataset with $\mathcal{G} = \{G_1,\dots,G_N\}$.

\paragraph{Graph Counterfactuals}
Given a black-box (oracle) predictor $\Phi: G \rightarrow Y$, where, w.l.o.g., $Y = \{0,1\}$, according to \cite{prado2022survey}, a counterfactual for $G$ is defined as
\begin{equation}\label{eq:gce_counterfactuality}
\mathcal{E}_{\Phi} \left( G \right) = \underset{G^{\prime} \in \mathcal{G}^{\prime}, G \neq G^{\prime}, \Phi \left( G \right) \neq \Phi \left( G^{\prime} \right)}{\arg\max} \mathcal{S} \left( G, G^{\prime} \right)
\end{equation}
where $\mathcal{G}'$ is the set of all possible counterfactuals, and $\mathcal{S}(G,G')$ calculates the similarity between $G$ and $G'$. \cite{prenkaj2023dygrace} reformulate Eq. \ref{eq:gce_counterfactuality} and take a probabilistic perspective to produce a counterfactual that is quite likely within the distribution of valid counterfactuals by maximising
\begin{equation}\label{eq:prenkaj_counterfactuality}
    \mathcal{E}_{\Phi}(G) = \underset{G' \in \mathcal{G}'}{\arg \max}\; P_{cf} \left( G' \; \middle| \; G, \Phi \left( G \right), \neg \Phi\left( G \right) \right)
\end{equation}
where $\neg \Phi(G)$ indicates any other\footnote{The provided formulation supports multi-class classification problems. For simplicity, we are in binary classification.} class from the one predicted for $G$. In this work, we find valid counterfactuals by solving a specialisation of Eq. \ref{eq:prenkaj_counterfactuality}.

\section{Method}\label{sec:method}

\begin{figure*}[!t]
    \centering
    \includegraphics[width=.9\textwidth]{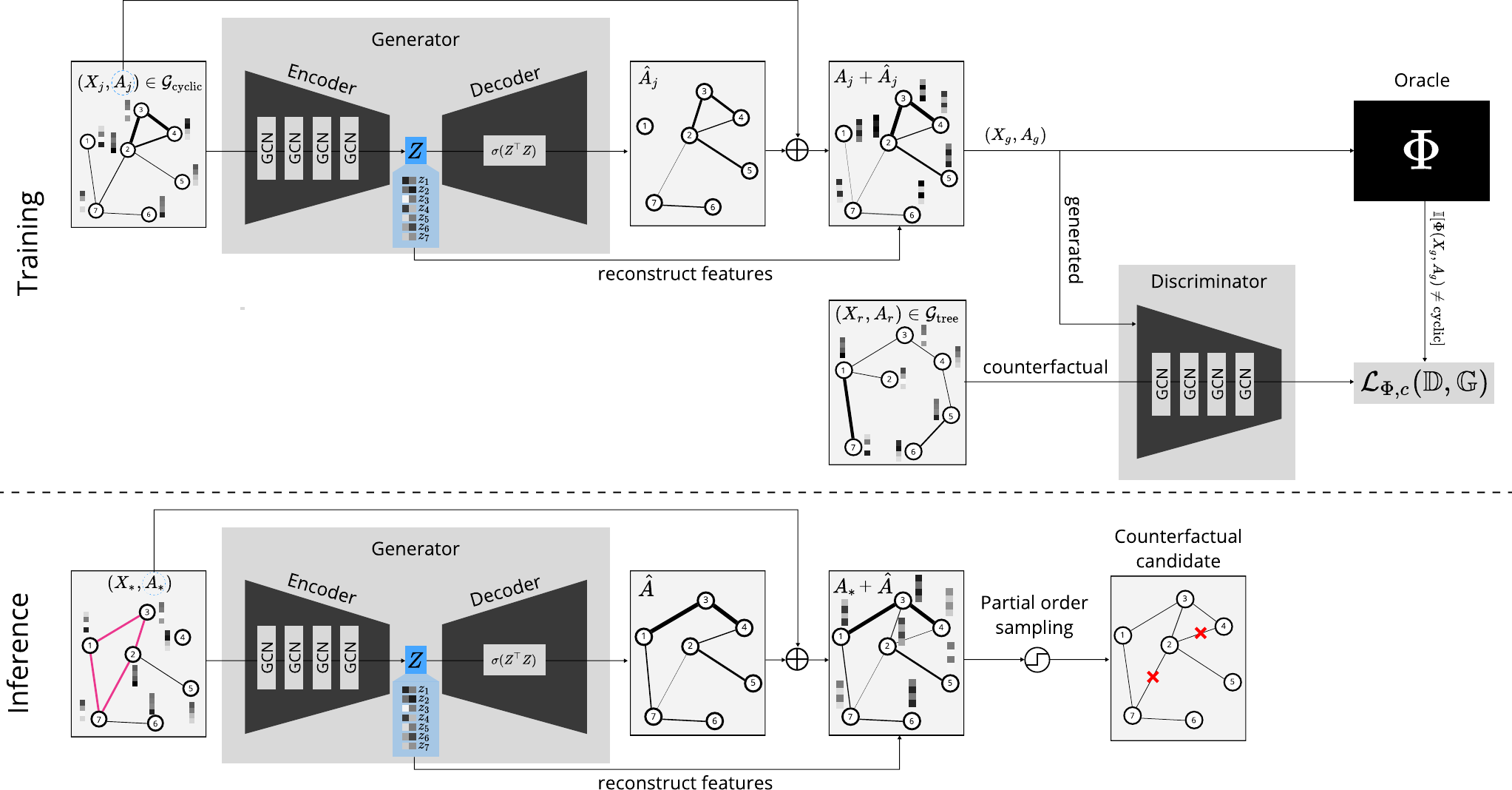}
    \caption{\OUR's workflow during training (up) and inference (down) in a cyclic graph vs tree scenario. $(X_j,A_j) \in \mathcal{G}_{\text{cyclic}}$ is fed to the generator, which produces counterfactual residuals. The discriminator is trained on both real $(X_r,A_r) \in \mathcal{G}_{\text{tree}}$ and generated graphs $(X_g,A_g)$. The generator optimisation is guided via $\Phi$'s classification. The generator explains input instances $(X_*,A_*)$ at inference time by sampling edges according to a learned stochastic distribution of its latent space. In this scenario, we highlight the critical edges forming a cycle the generator needs to break so that the returned counterfactual candidate does not contain cycles. The sampled edges entail a counterfactual candidate that $\Phi$ needs to validate.}
    \label{fig:method}
\end{figure*}

Here we introduce \OUR, namely \LONGOUR, a supervised explanation method that exploits the generator's learned latent space to sample plausible counterfactuals for a particular explainee graph $G =(X,A)$. \OUR distinguishes between the training and inference phases as depicted in Figure \ref{fig:method}. For this reason, we first discuss the training phase and, later, the inference one.

\paragraph{Training} Let $\mathbb{D}(Y| X,A)$ denote a GCN discriminator that produces $Y = \{0,1\}$ based on the plausibility assessment of the graph represented by the adjacency matrix $A$ and node features $X$. Let $\mathbb{G}$ represent a Graph Autoencoder (GAE) \cite{kipf2016variational} with $\text{ENC}(X,A)$ serving as the representation model for latent node-related interaction components (i.e., encoder), and $\text{DEC}(Z)$ acting as an inner-product decoder. Conversely to the original Residual GAN, \cite{zhang2020ris}, where the generator $\mathbb{G}$ is trained on sampled Gaussian noise, we exploit the Residual GAN introduced in \cite{nemirovsky22}, where $\mathbb{G}$ is trained on instances sampled from the data distribution and optimised according to Eq. \ref{eq:res_gan}:
\begin{equation}\label{eq:res_gan}
\begin{gathered}
    \mathcal{L}(\mathbb{D}, \mathbb{G}) = \underset{\begin{subarray}{c}(X_i,A_i)\in\mathcal{G}\end{subarray}}{\mathbb{E}} \underbrace{\bigg[\log \mathbb{D}(Y\mid X_i,A_i)\bigg]}_{\text{discriminator optimisation}}\\
    + \underset{\begin{subarray}{c}(X_j, A_j)\in \mathcal{G},\\\hat{X}_j , A_j + \hat{A}_j = \mathbb{G}(X_j,A_j)
    \end{subarray}}{\mathbb{E}} \underbrace{\bigg[\log (1 - \mathbb{D}(Y \mid \hat{X}_j, A_j + \hat{A}_j))\bigg]}_{\text{generator optimisation}}
\end{gathered}
\end{equation}
where $\mathbb{G}(X,A) = \text{DEC}(\text{ENC}(X,A))$. For completeness purposes, the generator $\mathbb{G}$ returns the reconstructed adjacency tensor $\hat{A}$ and the reconstructed node features $\hat{X}$. Notice that, unlike vanilla GANs, the input to the Residual GAN's discriminator is $A_j + \hat{A}_j$ for a graph $(X_j,A_j) \in \mathcal{G}$. To overcome the constraint of \cite{nemirovsky22}, which fixes the latent space of the generator to be of the same size as the input space, we exploit a Graph Autoencoder (GAE) such that it encodes graphs into a point in the latent space $\mathcal{Z}$,  and the decoder maps it into a new reconstructed graph belonging to the same space as the input. This also allows the generator to learn relationships between its input and output, enabling fine-grained regularisation of residuals and alleviating mode collapse. Moreover, to support edge additions and removals, we set the activation function of the generator to the hyperbolic tangent and sum it to the input adjacency matrix.

Let $\mathbb{I}[\Phi(X,A) \neq c]$ be an indicator function that returns 1 if $\Phi(X,A) \neq c$ for a graph $G=(X,A)$. Let also be $\mathcal{G}_c = \{(X,A)\;|\; (X,A) \in \mathcal{G} \; \land \neg \mathbb{I}[\Phi(X,A) \neq c]\}$. Conversely, we indicate with $\mathcal{G}_{\neg c} = \{(X,A)\;|\; (X,A) \in \mathcal{G} \; \land \mathbb{I}[\Phi(X,A) \neq c]\}$. Because we need to generate counterfactuals for a pre-trained black-box oracle $\Phi$ in a particular class $c$, we modify Eq. \ref{eq:res_gan} as follows:
\begin{equation}\label{eq:res_gan_oracle}
\begin{gathered}
    {\mathcal{L}}_{\Phi, c}(\mathbb{D},\mathbb{G}) = \sum_{\begin{subarray}{c}(X_r,A_r) \in \mathcal{G}_{\neg c}\end{subarray}}\underbrace{\bigg( \log \mathbb{D}(Y\mid X_r,A_r)\bigg)}_{\text{discriminator optimisation on real data}}\\    +     \sum_{\begin{subarray}{c}(X_g,A_g) \in  \mathbb{G}(\mathcal{G}_{c})   \end{subarray}}\underbrace{\bigg( \mathbb{I}[\Phi(X_g,A_g) \neq c]\log \mathbb{D}(Y\mid X_g,A_g)\bigg)}_{\text{discriminator optimisation on generated data}}\\    + \sum_{\begin{subarray}{c}
        (X_j,A_j) \in \mathcal{G}_c,\\
        \hat{X}_j,A_j + \hat{A}_j = \mathbb{G}(X_j,A_j)
    \end{subarray}} \underbrace{\log\bigg(1-\mathbb{D}(Y \mid \hat{X},A_j+\hat{A}_j)\bigg)}_{\text{generator optimisation}}\\
\end{gathered}
\end{equation}
where $\mathbb{G}(\mathcal{G}_{c})= \{\mathbb{G}(X,A)\;|\; (X,A) \in \mathcal{G} \; \land  \neg \mathbb{I}[\Phi(X,A) \neq c]\}$ is the set of all the generated graphs. %where $\mathbb{I}[\Phi(X,A) = c]$ is an indicator function that returns 1 if $\Phi$ classifies as factual the considered graph $G=(X,A)$ to penalise the generator in case it misproduces a factual. 
Since sampling instances from the data distribution might induce $\mathbb{G}$ to generate null residuals, integrating the accuracy of correct predictions from $\Phi$ - as shown in the second summation of Eq. \ref{eq:res_gan_oracle} - steers the generator away from this behaviour, making it produce realistic counterfactuals \cite{guyomard2022vcnet} w.r.t. to an input graph $G_* = (X_*,A_*)$, i.e.,
\begin{equation}\label{eq:realistic_counterfactuals}
    \underset{\begin{subarray}{c}(X,A) \in \mathcal{G},\\ \Phi(X,A) = \Phi(X_*, A_*)\end{subarray}}{\mathbb{E}}\bigg[\big|\big|\mathbb{G}(X_*,A_*)-(X,A)\big|\big|^2_2\bigg]
\end{equation}
where $G = (X, A)$ is a graph belonging to the same class as $G_*$. In other words, our approach produces counterfactuals close to the examples the generator has been trained, thus ensuring as little as possible perturbations w.r.t. $G_*$ (see Sec. \ref{sec:experiments}).

We train the generator exclusively on the graphs from the class we aim to explain. In this way, the graphs from the other classes are assigned as real instances for the discriminator training. Hence, by training the discriminator to differentiate between fake data (generated counterfactuals) and real data (corresponding to true counterfactual classes), the generator learns to produce counterfactuals conditioned on the graphs from the explainee class. For example, suppose we have a dataset containing acyclic (0) and cyclic (1) graphs. If we want to generate counterfactuals for acyclic, we feed the generator with acyclic instances. Meanwhile, we feed the discriminator with generated cyclic (labelled as fake data) and real cyclic (labelled as real data). Considering that the generator needs to fool the discriminator to maximise its objective function, it will learn how to mutate an  acyclic into a plausible cyclic (e.g., by adding a single edge to form a cycle). This way, the generator's latent space can be exploited to generate counterfactual candidates for the input instance. 

RSGG-CE can also be adapted for node classification where the dataset, the oracle, and the optimisation function (Eq. \ref{eq:res_gan_oracle}) must be modified accordingly. We invite the reader to check Sec. \ref{sec:node_adapation} for further details.

\paragraph{Inference and partial order sampling} Differently from
\cite{ma2022clear}, we stochastically generate counterfactual candidates by sampling edges with partial order guided by the learned probabilities from the generator's latent space.
The general sampling procedure is illustrated in Algorithm \ref{algo:sampling}, where we sample the edges according to the order defined by the \texttt{partial\_order} (line 3) function illustrated in Algorithm \ref{algo:partial_order}. 
We invite the reader to note that this approach is modular and that the function \texttt{partial\_order($\cdot$,$\cdot$)} can be specialised according to the application domain and prediction scenario.
In Algorithm \ref{algo:sampling} for each edge set $\mathcal{E}$ of each partition group $\mathbb{O}$, we sample each edge according to their estimated probabilities and engender $A^\prime$ (line 7) and assess if it is a valid counterfactual (line 8) based on the verification guard $o$. As in \cite{abrate2021counterfactual}, we default to the original input if we fail to produce a valid counterfactual (line 13).
In Algorithm \ref{algo:partial_order}, we induce a partial order on the estimated edges, including non-existing ones in the original input.   
In line 1, we get the edges of the input instance, while in line 2, we get the set of non-existing ones. Afterwards, in line 3, we build the corresponding partition groups by setting the oracle verification guard $o=1$ only for the non-existing group. Thus, the oracle will be called only once sampling finishes on the existing edge set.

\begin{algorithm}[!t]
\caption{Partial order sampling to produce a counterfactual.}\label{algo:sampling}
\begin{algorithmic}[1]
\Require$G_* = (X_*,A_*)$, $\mathbb{G}:\mathcal{G} \rightarrow \mathcal{G}$, $\Phi$, 
\State $\hat{X}_*,  A_* + \hat{A}_* = \mathbb{G}(X_*,A_*)$
\State $X_g, A_g \gets \hat{X}_*, A_* + \hat{A}_*$
\State $\mathcal{P} \gets \text{\texttt{partial\_order}}(A_*)$ 
\State $A^\prime \gets 0^{n\times n}$
\For{$ \mathbb{O} \in \mathcal{P}$} 
    \For{$e = (u,v) \in \mathbb{O}.\mathcal{E}$}
        \State $A^\prime[u,v] \gets \text{\texttt{sample}}(e, A_g[u,v])$
        \If{$\mathbb{O}.o \land \Phi(X_g, A^\prime) \neq \Phi(X_*,A_*)$}
    \State \Return $(X_g, A^\prime)$
    \EndIf
    \EndFor
\EndFor
\State \Return $(X_*,A_*)$
\end{algorithmic}
\end{algorithm}

\begin{algorithm}[!t]
\caption{Example of \texttt{partial\_order}}\label{algo:partial_order}
\begin{algorithmic}[1]
\Require $A \in \mathbb{R}^{n \times n}$ 
\State $E \gets \text{\texttt{positive\_edges}}(A)$\Comment{\textit{Get the set of edges from the adjacency matrix $A$}}
\State $\neg E \gets \text{\texttt{negative\_edges}}(A)$\Comment{\textit{Get the set of non-existing edges from the adjacency matrix $A$}}
\setlength{\thickmuskip}{0mu}
\State $\mathcal{P} \gets \{(\mathcal{E}=E,o=0) , (\mathcal{E}=\neg E,o=1)\}$\Comment{\textit{Build the partial order of the existing and non-existing edges with group tuples consisting of edge set $\mathcal{E}$,  and oracle verification guard $o$.}}
\State \Return $\mathcal{P}$
\end{algorithmic}
\end{algorithm}

\section{Experimental Analysis}\label{sec:experiments}

\begin{table}[!t]
\centering
\caption{Comparison of \OUR with SoA methods. Metrics are reported on 10-fold cross-validations. Bold values are the best overall; underlined are second-best; $\times$ represents no convergence; $\dag$ depicts a learning-based explainer, and $\ddag$ a generative approach. The highlighted rows represent the most important metrics.}
\label{tab:soa_performance}
\resizebox{\linewidth}{!}{%
\begin{tabular}{@{}llccccc@{}}
\toprule
\multicolumn{2}{l}{}                                              & \multicolumn{5}{c}{Methods}                                                      \\ \cmidrule(l){3-7} 
\multicolumn{2}{l}{\multirow{-2}{*}{}} &
  MEG $\dag$ &
  CF$^2$ $\dag$ &
  CLEAR $\ddag$ &
  G-CounteRGAN $\ddag$ &
  \textbf{RSGG-CE $\ddag$} \\ \midrule
                                      & Runtime (s) $\downarrow$  & 272.110        & \underline{4.811}   & 25.151   & 632.542  & \textbf{0.083}      \\
                                      & GED $\downarrow$          & 159.700        & \underline{27.564}  & 61.686   & 182.414  & \textbf{11.000}     \\
                                      & Oracle Calls $\downarrow$ & \textbf{0.000} & \textbf{0.000}      & 4341.600 & 1321.000 & \underline{121.660} \\
 &
  \cellcolor[HTML]{EFEFEF}Correctness $\uparrow$ &
  \cellcolor[HTML]{EFEFEF}\underline{0.530} &
  \cellcolor[HTML]{EFEFEF}0.496 &
  \cellcolor[HTML]{EFEFEF}0.504 &
  \cellcolor[HTML]{EFEFEF}0.504 &
  \cellcolor[HTML]{EFEFEF}\textbf{0.885} \\
                                      & Sparsity $\downarrow$     & 2.510          & 0.496               & 1.110    & 3.283    & \textbf{0.199}      \\
 &
  \cellcolor[HTML]{EFEFEF}Fidelity $\uparrow$ &
  \cellcolor[HTML]{EFEFEF}\underline{0.530} &
  \cellcolor[HTML]{EFEFEF}0.496 &
  \cellcolor[HTML]{EFEFEF}0.504 &
  \cellcolor[HTML]{EFEFEF}0.504 &
  \cellcolor[HTML]{EFEFEF}\textbf{0.885} \\
\multirow{-7}{*}{\rotatebox{90}{TC}}  & Oracle Acc. $\uparrow$    & 1.000          & 1.000               & 1.000    & 1.000    & 1.000               \\ \midrule
                                      & Runtime (s) $\downarrow$  & $\times$       & \textbf{15.313}     & 275.884  & 969.255  & \underline{80.000}  \\
                                      & GED $\downarrow$          & $\times$       & \underline{655.661} & 1479.114 & 3183.729 & \textbf{234.853}    \\
                                      & Oracle Calls $\downarrow$ & $\times$       & \textbf{0.000}      & 5339.455 & 1182.818 & \underline{794.805} \\
 &
  \cellcolor[HTML]{EFEFEF}Correctness $\uparrow$ &
  \cellcolor[HTML]{EFEFEF}$\times$ &
  \cellcolor[HTML]{EFEFEF}0.463 &
  \cellcolor[HTML]{EFEFEF}\underline{0.554} &
  \cellcolor[HTML]{EFEFEF}0.529 &
  \cellcolor[HTML]{EFEFEF}\textbf{0.603} \\
                                      & Sparsity $\downarrow$     & $\times$       & \underline{0.850}   & 1.917    & 4.125    & \textbf{0.304}      \\
 &
  \cellcolor[HTML]{EFEFEF}Fidelity $\uparrow$ &
  \cellcolor[HTML]{EFEFEF}$\times$ &
  \cellcolor[HTML]{EFEFEF}\textbf{0.287} &
  \cellcolor[HTML]{EFEFEF}\underline{0.319} &
  \cellcolor[HTML]{EFEFEF}0.265 &
  \cellcolor[HTML]{EFEFEF}\textbf{0.287} \\
\multirow{-7}{*}{\rotatebox{90}{ASD}} & Oracle Acc. $\uparrow$    & $\times$       & 0.773               & 0.773    & 0.773    & 0.773               \\ \bottomrule
\end{tabular}%
}
\end{table}

In this section, we discuss three kinds of analyses\footnote{The original code is on \textcolor{blue}{\url{https://github.com/MarioTheOne/GRETEL}}. Newer versions of the project will be released on \textcolor{blue}{\url{https://github.com/aiim-research/GRETEL}}.}. First, we discuss the performances of \OUR w.r.t. other SoA explainers (see Sec. \ref{appendix:eval_metrics} for the used evaluation metrics). In the second, we conduct four ablation studies to understand the robustness of \OUR. 
In the third one, we conduct a qualitative anecdotal inspection. 
Experimental analysis is done by applying 10-fold cross-validations on real and synthetic datasets.  We also analysed \OUR's efficiency and convergence in Sec. \ref{appendix:eff_conv}.

The \textit{Tree-Cycles (TC)} \cite{ying2019gnnexplainer}  is an emblematic synthetic dataset. 
Each instance constitutes a graph comprising a central tree motif and multiple cycle motifs connected through singular edges. 
The dataset encompasses two distinct classes: i.e., one for graphs without cycles (0) and another for graphs containing cycles (1). The TC also allows control of the number of nodes, the number of cycles and the number of nodes in them. 
For the performance comparison to the other SoA explainers, due to the computational complexity of some of them, we use $500$ graphs with $28$ nodes and randomly generate up to 3 cycles with varying sizes that go up to $7$ nodes. We vary all the parameters for the ablation study as reported in Sec. \ref{sec:ablation}. The hardness of this dataset depends on changing the oracle prediction; the explainer needs to learn to apply two opposite actions (i.e., remove or add edges to the input to generate acyclic or cyclic counterfactuals, respectively).

The \textit{Autism Spectrum Disorder} (\textit{ASD}) \cite{abrate2021counterfactual} is a real graph classification dataset obtained using functional magnetic resonance imaging (fMRI), where nodes represent brain Regions of Interest (ROI), and edges are co-activation between two ROIs. The two classes belong to individuals with \textit{Autism Spectrum Disorder} (ASD) and \textit{Typically Developed} (TD) individuals as the control group. Here, the graph instances can be disconnected.

In Table \ref{tab:soa_performance}, we report the performance of \OUR compared to other SoA learning-based explanation methods (i.e., 
MACCS \cite{wellawatte2022model}, CF$^2$ \cite{tancf2}, CLEAR \cite{ma2022clear}, and G-CounteRGAN  \cite{prado2023revisiting}) for the TC and ASD datasets.
Notably, a significant challenge arises when it comes to counterfactual explainability through learning-based explainers due to the inherent reliance on the overarching structure of the graph rather than specific nodes or edges. 
In this context, \OUR takes the spotlight as an exceptional performer with a gain of $66.98\%$ and $8.84\%$ in Correctness over the second-performing method in TC and ASD, respectively. 
It outperforms all alternative methods, standing as the sole technique that improves, by a large margin, the Correctness of all the datasets without sacrificing the running time.
Moreover, \OUR also showcases superiority in terms of Graph Edit distance (GED) w.r.t. the other explainers. 
This further underscores \OUR's capabilities in capturing the intricate structures in all the datasets and its wanted ability (see Sec. \ref{sec:prelimniaries} and Eq. \ref{eq:res_gan_oracle}) to generate counterfactual instances closer to the input one w.r.t. those generated by other explainers (see also Sec. \ref{sec:anecdotical}).

\subsection{Ablation Experiments}\label{sec:ablation}
To understand the robustness of \OUR, we conduct four ablation studies using the TC dataset by varying the number of nodes in the cycles, the number of cycles, the number of nodes in the graphs, and the number of instances in the dataset.
For all the ablations, we use a dataset with $500$ instances. 
We also include CF$^2$ since it is the best-performing learning-based explainer after \OUR.

\begin{figure}[!t]
    \centering
\includegraphics[width=.9\linewidth]{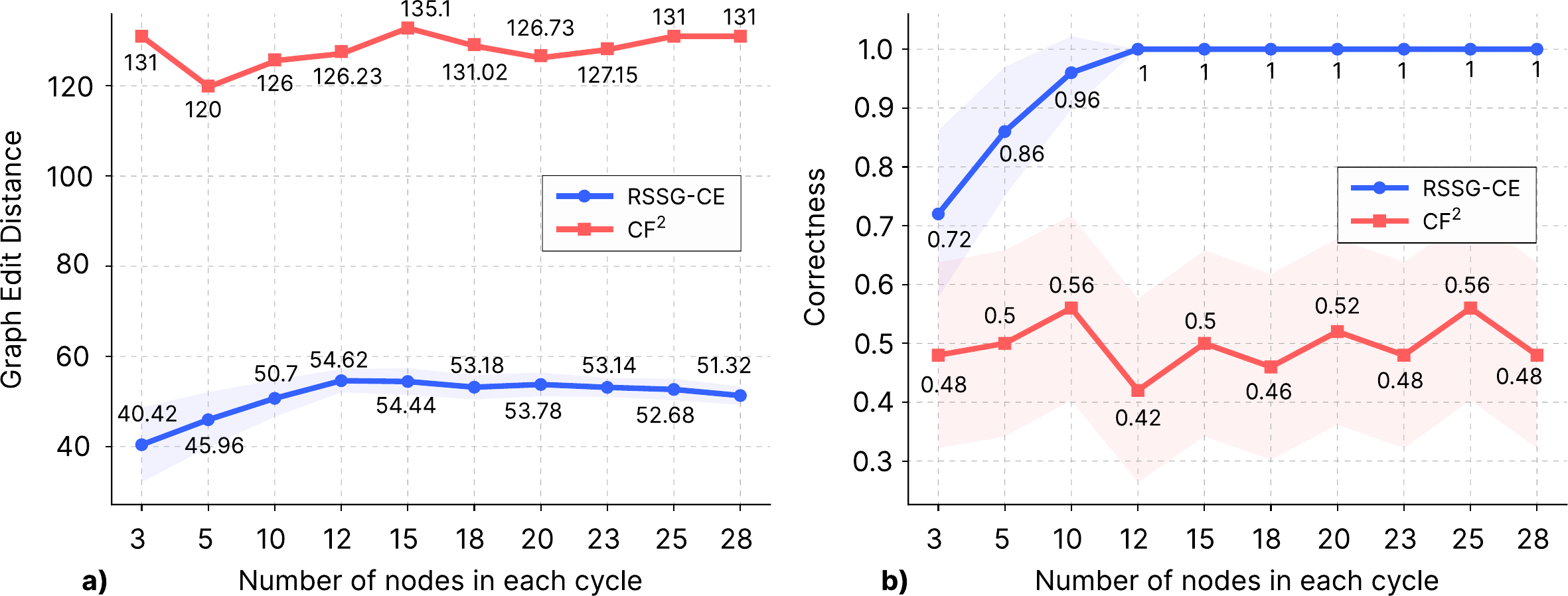}
    \caption{GED $\downarrow$ and Correctness $\uparrow$ trends when varying the number of nodes in each cycle on TreeCycles with 128 nodes and 4 cycles per instance.}
    \label{fig:ablation_on_cycle_size}
\end{figure}

\begin{figure}[!t]
    \centering
\includegraphics[width=.9\linewidth]{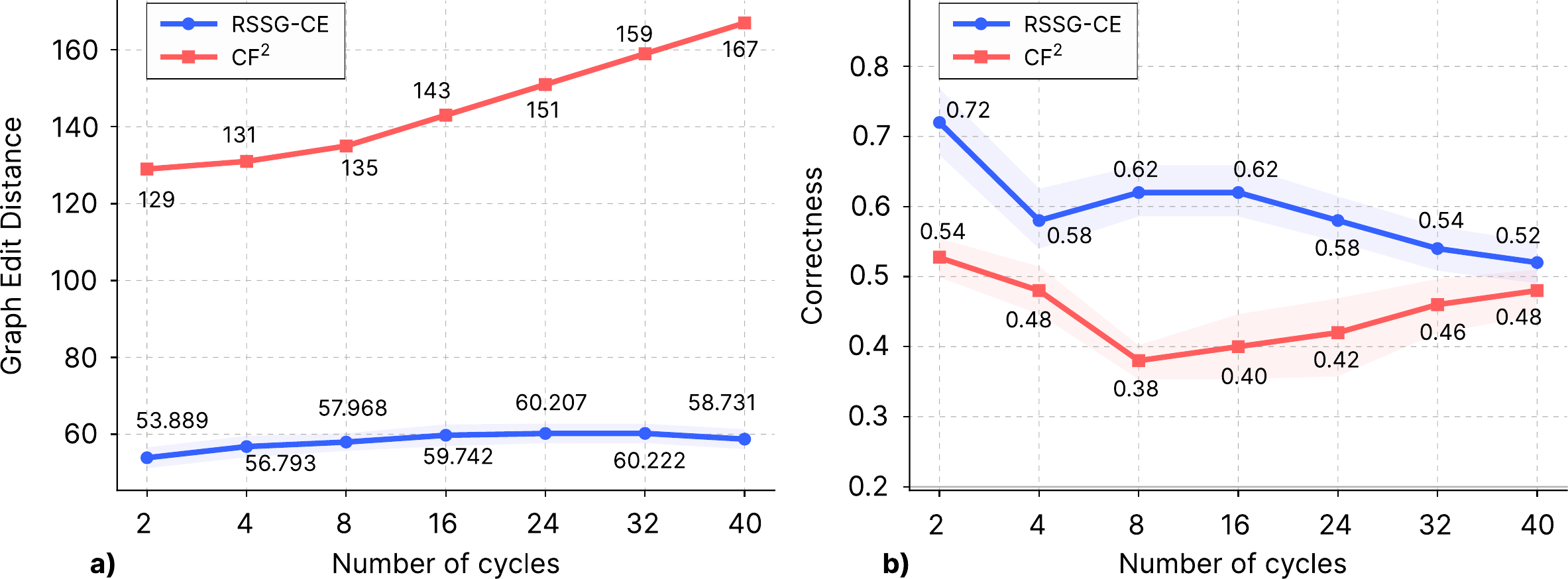}
    \caption{GED $\downarrow$ and Correctness $\uparrow$ trends when varying the number of cycles per instance on TreeCycles with 128 nodes and 3 nodes per cycle.}
    \label{fig:ablation_on_number_of_cycles}
\end{figure}

\paragraph{Robustness to the increasing number of nodes per cycle} 
In this study, we fix the number of nodes to $128$ and the number of cycles to $4$ to assess how the number of nodes (from $3$ to $28$) in each cycle affects \OUR's GED and Correctness (see Fig. \ref{fig:ablation_on_cycle_size}). 
It is interesting to notice that the bigger the cycles become, the better \OUR learns the edge probabilities since now the motifs are more evident.
This means that, at inference time, the partial order sampling has higher chances of breaking cycles containing more nodes than those with a small number of them. For instance, if the number of nodes in a cycle is $3$, the probability of cutting the cycle is $2/3$. 
Specularly, if the number of nodes in a cycle tends to the number of nodes in the instance, the probability of cutting the cycle is nearer to $1$. 
Therefore, Correctness has an increasing trend. 
Additionally, while \OUR reaches a correctness of 1 when the number of nodes in cycles increases, it does not sacrifice the recourse cost (GED) needs to engender valid counterfactuals (see Fig. \ref{fig:ablation_on_cycle_size}.a). 
Contrarily, since \OUR's partial order sampling favours the existing edges in the original instance, with the increase in cycle sizes, the GED has a non-increasing trend. 
This is straightforward since finding a valid counterfactual can be done with fewer sampling iterations. 
For example, if we have a single ring (cyclic graph) per instance, the number of operations to engender a counterfactual is $1$. 
Notice that CF$^2$ entails a zig-zag Correctness trend, leading us to believe it does not scale with the increasing number of  nodes in the cycle. 
This phenomenon is also supported regarding GED since CF$^2$ converges to a local minimum and tries to engender the same counterfactual regardless of the input instance (notice the standard error equal to zero). 

\begin{figure}[!t]
    \centering
\includegraphics[width=.9\linewidth]{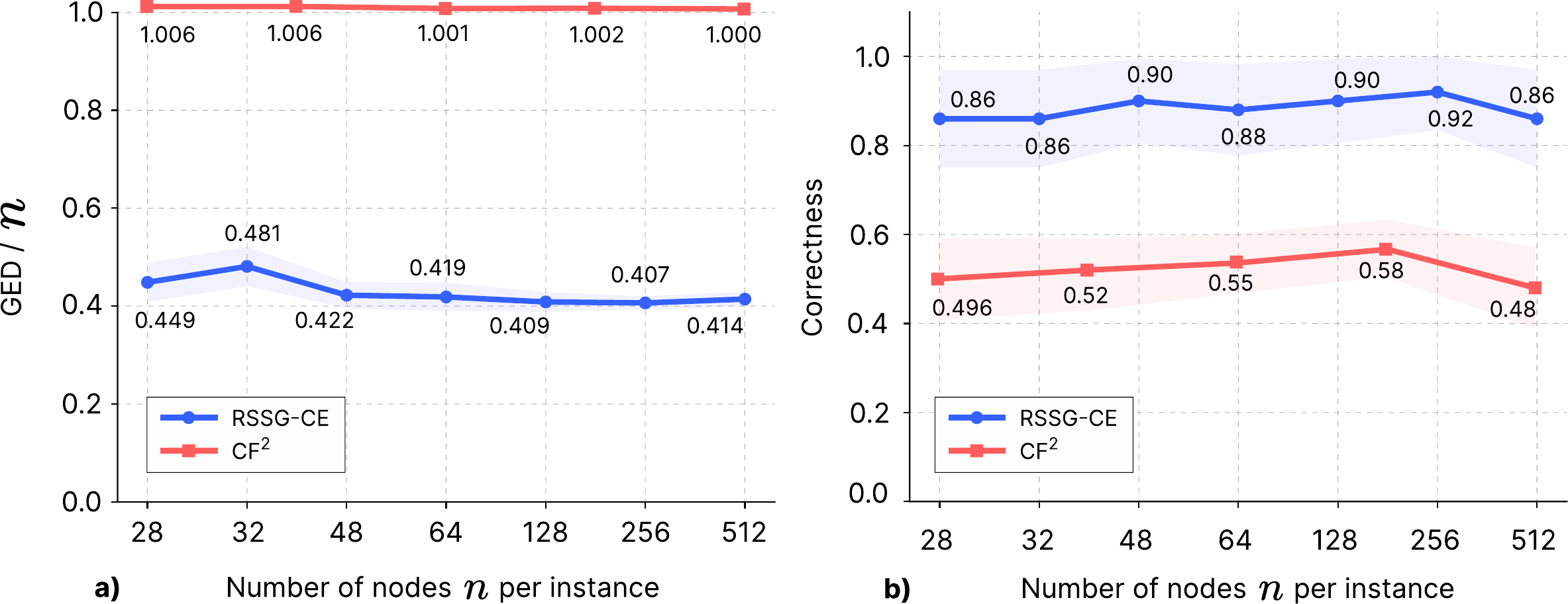}
    \caption{The trend of GED $\downarrow$ normalised by the number $n$ of nodes instance, and the trend of Correctness $\uparrow$ when $n$  increases.}\label{fig:ablation_number_of_nodes_per_instance}
\end{figure}

\begin{figure}[!t]
    \centering
\includegraphics[width=.9\linewidth]{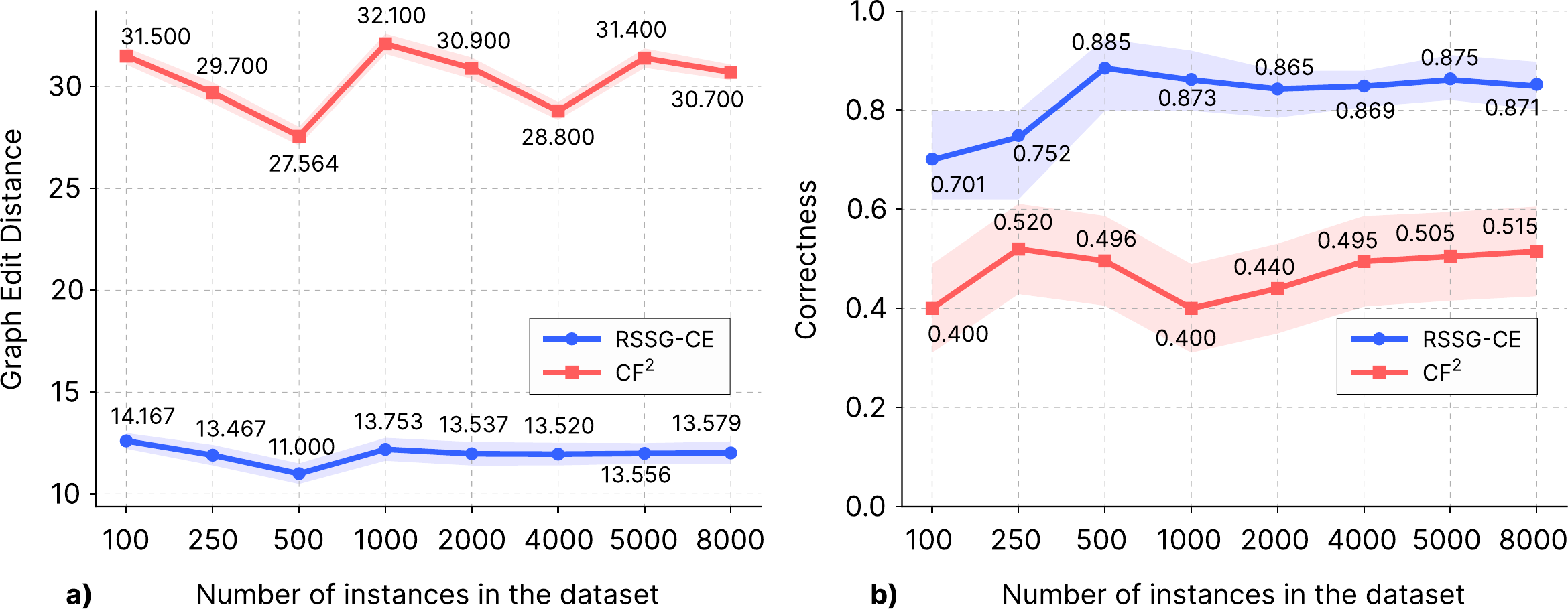}
    \caption{The trend of GED $\downarrow$ and Correctness $\uparrow$ when varying the number of instances in the dataset.}
    \label{fig:ablation_dataset_size}
\end{figure}

%%%%%%
\begin{figure*}[!t]
    \centering
    \includegraphics[width=0.8\textwidth]{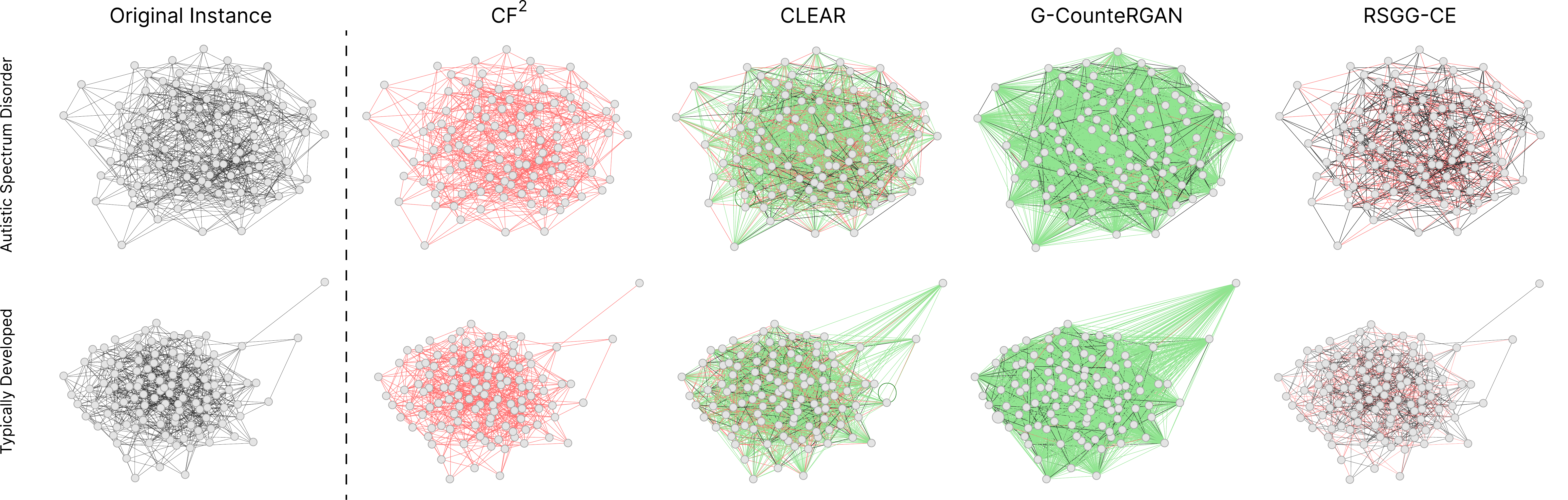}
    \caption{Qualitative comparison (best viewed in colour) of the counterfactuals candidate produced by CF$^2$, CLEAR, G-CounteRGAN, and \OUR on the ASD real dataset. Concerning the original graph, green edges are additions, red are removals, and grey are maintained.}
    \label{fig:qualitative_exp}
\end{figure*}
%%%%%%

\paragraph{Robustness to the increasing number of cycles}
In this investigation, we keep the cycle size fixed at $3$ nodes and assess the impact of cycle quantity (ranging from $2$ to $32$) on \OUR's GED and Correctness metrics (Fig. \ref{fig:ablation_on_number_of_cycles}). 
It is interesting to notice that the GED (Fig. \ref{fig:ablation_on_number_of_cycles}.a) is linearly dependent on the number of cycles, thus reflecting the increased hardness of the problem (i.e., the need to break a higher number of cycles) but without slipping into an exponential trend. Similarly, the Correctness is slightly affected by the number of cycles in the graphs but maintains its linear trend. This plot indicates that \OUR can cope with complex topologies without sacrificing too much in terms of performance. Additionally, CF$^2$ has an increasing trend in GED w.r.t. the change in the number of cycles also influences the Correctness (e.g., from $2$ to $18$).

\paragraph{Robustness to the number of nodes}
In this ablation,  for each graph, we randomly generated up to $3$ cycles with a varying size that goes up to $7$ nodes, and we delve into the impact of graph number of nodes (ranging from $28$ to $512$) on \OUR's GED and Correctness metrics (refer to Fig. \ref{fig:ablation_on_number_of_cycles}).
To better catch the general trend,  in this case, we reported the recourse cost (GED) normalised by the number of nodes of the graph. Fig. \ref{fig:ablation_number_of_nodes_per_instance}.a. Concerning almost constant value, which depicts an outstanding linear trend w.r.t. the number of nodes of the graph. It must be noticed that the Correctness - reported in Figure \ref{fig:ablation_number_of_nodes_per_instance}.b is not affected by the increased dimensionality.
Those results, in conjunction with the results obtained in the previous ablation study (see Fig. \ref{fig:ablation_on_number_of_cycles}.b), let us state that the performances are solely affected by the complexity of the datasets. Similarly, CF$^2$ has a linear trend in normalised GED by $n$ as \OUR. Recall that CF$^2$ is a factual-based explainer that supports only edge removal operations. Now, because the GED / $n$ ratio tends to $1$, we argue that CF$^2$ perturbs the entire adjacency matrix of the original instance.

\paragraph{Robustness by the number of instances} In this last ablation,  we randomly generated up to 3 cycles with varying sizes that go up to 7 nodes for each graph. We delve into the impact of varying the number of instances in the dataset (ranging from 100 to 8000) on \OUR's GED and Correctness metrics (refer to Fig. \ref{fig:ablation_dataset_size}). Notice that in Fig. \ref{fig:ablation_dataset_size}.a, the GED is unaffected by the number of instances by exposing a linear constant trend for both \OUR and CF2. As expected, the Correctness of \OUR increases at the beginning and stabilises when the number of instances exceeds 250 (see Fig. \ref{fig:ablation_dataset_size}.b). In general, the same Correctness trend is also confirmed for CF$^2$ with a lower value.

\subsection{Qualitative Anecdotal Inspection}\label{sec:anecdotical}

Here, we discuss anecdotally the quality of the counterfactual graphs generated by \OUR and the SoA methods on the ASD dataset.
Fig. \ref{fig:qualitative_exp} shows the counterfactual generation for CF$^2$, CLEAR, G-CounteRGAN, and \OUR on two graphs belonging to the \textit{Autistic Spectrum Disorder} and \textit{Typically Developed} classes.  For both instances, we show the original edges and illustrate how they get modified by each method. 
For visualisation purposes, we colour red the edge deletion operations, green the edge addition operations, and grey the original edge maintenance operations. CF$^2$ clearly shows its behaviour of removing the factual subgraph in both instances regardless of the class on ASD. However, it is peculiar that the subgraph $\tilde{G}$ corresponds to the original graph $G$, justifying the high GED reported in Table \ref{tab:soa_performance}. 
CLEAR is the only SoA method that, in this example, evidently performs all three types of operations on the original edges. Although not as naive as CF$^2$'s edge perturbation policy, CLEAR exposes a higher GED than CF$^2$ and \OUR due to its tendency to over-generate non-existing edges. Interestingly, CLEAR is the only method that estimates self-loops (see instance of class \textit{Typically Developed}) among the rest. G-CounteRGAN, as shown in Table \ref{tab:soa_performance}, is the worst-performing strategy, exposing a tendency to produce a densely connected graph by adding non-existing edges. Lastly, \OUR, although capable of edge addition/removal operations, here exhibits only removal ones. We argue that because \OUR's produced valid explanation has a low GED, this dataset's partial order sampling strategy engenders a valid counterfactual only by examining the original edges. To this end, we believe that edge removals are primarily needed to produce valid counterfactuals in ASD.  

We extend this qualitative analysis via a comprehensive visualisation technique that analyses the operations performed on each edge (see Sec. \ref{appendix:qualitative}).

\section{Conclusion}\label{sec:conclusion}
In this study, we introduced \OUR, a novel \LONGOUR able to produce counterfactual examples from the learned latent space considering a partially ordered generation sequence.   We showed, quantitatively and qualitatively, that \OUR outperforms all SoA methods. \OUR produces counterfactuals -- \textit{conditioned on the input} -- from the learned latent space incorporating a partially ordered generation sequence. Additionally, the proposed partial order sampling offers an effective means for discerning existing and non-existing edges, contributing to the overall robustness of our model. One of the key findings of our study is the resilience of \OUR in handling increasingly complex motifs within the graph instances. In the future, we want to leverage the ability of \OUR to generate multiple counterfactual candidates to  produce cohesive candidate explanations. Lastly, incorporating counterfactual minimality into the loss function of generative models might offer potential improvements in interpreting the generated explanations.

\section*{Acknowledgments}
This work is partially supported by the European Union - NextGenerationEU - National Recovery and Resilience Plan (Piano Nazionale di Ripresa e Resilienza, PNRR) - Project: ``SoBigData.it - Strengthening the Italian RI for Social Mining and Big Data Analytics'' - Prot. IR0000013 - Avviso n. 3264 del 28/12/2021, by the “ICSC – Centro Nazionale di Ricerca in High-Performance Computing, Big Data and Quantum Computing", funded by European Union – NextGenerationEU, by European Union - NextGenerationEU under the Italian Ministry of University and Research (MUR) National Innovation Ecosystem grant ECS00000041 - VITALITY - CUP E13C22001060006, and by Territori Aperti (a project funded by Fondo Territori, Lavoro e Conoscenza CGIL CISL UIL). 

\noindent All the numerical simulations have been realized on the Linux HPC cluster Caliban of the High-Performance Computing Laboratory of the Department of Information Engineering, Computer Science and Mathematics (DISIM) at the University of L’Aquila.

\bibliography{main}

\clearpage

\begin{appendix}

\noindent We enhance the main paper (MP) by incorporating an exhaustive description of the datasets utilised in the experiments. We thoroughly depict the hyperparameter search for our proposed method (\OUR) and SoA methods. We give the reader comprehensive instructions for replicating the precise outcomes demonstrated in MP. Additionally, we delve into an expanded discourse concerning the counterfactuals generated, supplemented by illustrative pictograms that encapsulate insights gleaned from 10-fold cross-validations.

\section{Datasets descriptions}\label{appendix:datasets}

To assess \OUR's performances, we used two datasets from existing literature: i.e., Tree-Cycles (TC) and Autism Spectrum Disorder (ASD). The former is artificially generated, affording us control over diverse parameters like instance count, nodes within each instance, and cycle quantity per instance. This aspect greatly facilitates the execution of ablation studies. Conversely, ASD is a compact, real dataset obtained from the biomedical sector, serving as a valuable means to evaluate explanation techniques within a genuine context. Table \ref{tab:gr_datasets} provides the reader with relevant statistical information about the two datasets.

\begin{table}[h]
\centering
\caption{Statistical description of the datasets.}
\label{tab:gr_datasets}
\resizebox{\linewidth}{!}{%
\begin{tabular}{@{}lrrr@{}}
\toprule
Dataset            & Tree-Cycles   & ASD  \\ \midrule
\# of instances    &  500          & 101   \\
Avg graph diameter & 12.812        & $\infty$ \\
Avg \# of nodes    & 28            & 116          \\
Avg \# of edges    & 27.566        & 655.624     \\
Max \# of nodes    & 28            & 116        \\
Avg node degree    & 1.969         & 11.304        \\ 
\# of classes      & 2             & 2         \\
Class distribution & 0.504 : 0.496 & 0.515 : 0.485        \\
Graph type         & Undirected    & Undirected \\
Connected components & One & Multiple\\
 \bottomrule
\end{tabular}%
}
\end{table}

\textit{Tree-Cycles} \cite{ying2019gnnexplainer,prado2022survey} is a dataset for graph classification, where each instance has a random tree as a base structure, cycle motifs are added to each instance, and the resulting graph is binary-classified based on the presence of a cycle. The user can control the number of instances, the number of vertices per instance, and the number of connecting edges. This method is similar to what \cite{ying2019gnnexplainer} proposed but with the addition of minimal counterfactual explainability ground truth for each instance. All the instances in the dataset contain the same number of nodes and are connected graphs. Notice that the dataset has a balanced distribution between its two classes, making it conducive for learning-based explainers. However, its average node degree of 1.969 suggests the graphs are sparsely connected, challenging GNNs to capture intricate relationships between the nodes.

Additionally, notice that this dataset poses a difficult scenario since explainers need to learn both edge additions/removal operations given the duality aspect of the instances. In other words, explainers need to learn how to remove edges to pass from a cyclic graph to an acyclic graph and how to add edges to pass from a tree to a cyclic graph. Although the dataset poses a difficult scenario for most learning-based approaches, as seen in Table \ref{tab:soa_performance}, \OUR's priority order sampling strategy enables us to effectively surpass the SoA with an improvement of $66.98\%$ in terms of Correctness (see also Fig. \ref{fig:pictorial_TC}).

\textit{Autism Spectrum Disorder} (\textit{ASD}) \cite{abrate2021counterfactual} is a graph classification dataset focused on children below nine years of age. The dataset encompasses 49 individuals with ASD, juxtaposed against 52 Typically Developed (TD) individuals forming the control group. Data extraction involves functional magnetic resonance imaging (fMRI), wherein vertices correspond to brain Regions of Interest (ROIs) and edges denote co-activation between two ROIs. Notably, all instances within this dataset exhibit the same number of nodes, with the majority being unconnected graphs. Despite the equal class distribution, the graphs exhibit a notably higher average node degree of 11.304, implying denser interconnectivity between nodes. This characteristic may present challenges in handling the complexity of interrelated features and distinguishing relevant patterns. The undefined average graph diameter ($\infty$) suggests considerable variability in graph sizes and structures due to multiple connected components, making it more intricate to discern consistent patterns across instances.

Moreover, multiple connected components in the graphs further complicate explanations, as they introduce varying degrees of separation between subgraphs. In other words, explainers would need to understand whether concentrating on the connected components singularly and perturbing them to generate counterfactual candidates or consider the graph as a whole and add/remove connections between the connected components (see Fig. \ref{fig:pictorial_ASD}). This structure diversity demands robust techniques for capturing and generalising meaningful graph representations that can generate valid counterfactuals.

\section{Hyperparameter search and optimisation}

In our experimental evaluation, we tested \OUR against other learning-based explanation methods from the literature. The competitors are CF$^2$ \cite{tancf2}, CLEAR \cite{ma2022clear}, and G-CounteRGAN \cite{prado2023revisiting}. To ensure a fair comparison, we performed a Bayesian optimisation to select the best hyperparameters for all SoA methods and \OUR with Correctness as the objective function. The hyperparameter optimisation was performed in the Tree-Cycles dataset because it constitutes the hardest scenario in the literature. We trained on an HPC SGE Cluster of 6 nodes with 360 cumulative cores, 1.2Tb of RAM, and two GPUs (i.e., one Nvidia A30 and one A100).  

% CF2 Hyperparameters
\begin{table}[h]

\caption{Hyperparameter search space and selection for CF$^2$}
\begin{tabular}{@{}lll@{}}
\toprule
 Hyperparameter & Search Space & Best Value \\ 
 \midrule
 $\alpha$         & $\{0.5, 0.6 \dots 1.0\}$     &  0.7    \\
 $\lambda$        & $\{20, 100, 500, 1000\}$     &  20     \\
 $\gamma$         & $\{0.1, 0.2 \dots 1.0\}$     &  0.9    \\
 lr    & $\{ 10^{-4}, 10^{-3}, 10^{-2}\}$     &  $10^{-2}$ \\
 epochs           & $\{50, 100, 200, 250, 500\}$ &  500    \\ 
 batch size ratio & $\{ 0.1, 0.15, 0.2 \}$       &  0.2    \\ 
 \bottomrule
\end{tabular}
\label{tab:cf2_hpo}
\end{table}

CF$^2$ \cite{tancf2} (Table \ref{tab:cf2_hpo}) was not able to significantly improve the Correctness and Graph Edit Distance of the explanations despite the hyperparameter optimisation. We argue that this is due to the factual-based nature of CF$^2$ explanations, which limits the method to edge removal operations only.

% CLEAR Hyperparameters
\begin{table}[h]
\centering
\caption{Hyperparameter search space and selection for CLEAR}
\begin{tabular}{@{}lll@{}}
\toprule
 Hyperparameter & Search Space & Best Value \\ 
 \midrule
 $\alpha$         & $\{0, 0.2, \dots, 1\}$         & $0.4$     \\
 $\lambda_{sim}$  & $[0.1,1.0]$     & $1.0$     \\
 $\lambda_{kl}$   & $[0.1,1.0]$     & $1.0$     \\
 $\lambda_{cfe}$   & $[0.1,1.0]$    & $0.1$     \\
 lr    & $\{ 10^{-4}, 10^{-3}, 10^{-2} \}$ & $10^{-2}$  \\
 epochs           & $\{ 100, 300, 600 \}$          & $600$     \\ 
 batch size ratio & $\{0.1, 0.15, 0.2\}$           & $0.15$    \\ 
 dropout          & $\{ 0.1, 0.25, 0.5 \}$         & $0.1$     \\ 
 \bottomrule
\end{tabular}
\label{tab:clear_hpo}
\end{table}

CLEAR \cite{ma2022clear} (Table \ref{tab:clear_hpo}) did not improve the Correctness of the explanations during the hyperparameter optimisation process. However, it reduced the Graph Edit Distance between the original instances and the counterfactuals of $\sim50\%$. One of the main challenges with this explanation method is its large amount of hyperparameters. As a result, we needed to reduce the search space for each to perform the search in a reasonable amount of time, similar to the other methods.

% Countergan Hyperparameters
\begin{table}[h]
\centering
\caption{Hyperparameter search space and selection for G-CounteRGAN}
\begin{tabular}{@{}lll@{}}
\toprule
 Hyperparameter & Search Space & Best Value \\ 
 \midrule
 epochs                        & $\{ 10, 20, \dots, 250 \}$       & $250$   \\ 
 batch size ratio              & $\{ 0.1, 0.15, 0.2 \}$           & $0.1$   \\ 
 \# generator steps     & $\{ 2, 3, 10, 30 \}$             & $2$     \\
 \# discriminator steps & $\{ 2, 3, 10, 30 \}$             & $3$     \\
 threshold     & $\{ 0.4, 0.5, 0.6, 0.7, 0.8 \}$  & $0.5$   \\
 \bottomrule
\end{tabular}
\label{tab:gcountergan_hpo}
\end{table}

G-CounteRGAN \cite{prado2023revisiting} (Table \ref{tab:gcountergan_hpo}) displayed only small improvements with the hyperparameter optimisation. Notice that the upper boundary (e.g., see epochs) of its search space is lower w.r.t. the other methods since it is more resource-expensive than the other methods, taking multiple days to complete a single step in the optimisation.

% RSGG-CE Hyperparameters
\begin{table}[h]
\centering
\caption{Hyperparameter search space and selection for RSGG-CE}
\begin{tabular}{@{}lll@{}}
\toprule
 Hyperparameter & Search Space & Best Value \\ 
 \midrule
 epochs                        & $\{ 50, 100, \dots, 500 \}$     & 500   \\ 
% sampling iterations           & $\{ 50, 100, \dots, 500 \}$     & 500   \\ 
 generator lr       & $\{ 10^{-4}, 10^{-3}, 10^{-2} \}$  & $10^{-3}$   \\
 discriminator lr   & $\{ 10^{-4}, 10^{-3}, 10^{-2} \}$  & $10^{-3}$   \\ 
 \bottomrule
\end{tabular}
\label{tab:rsgg_hpo}
\end{table}

\OUR (Table \ref{tab:rsgg_hpo}) benefitted from the hyperparameter optimisation more than other methods.  Although \OUR has a small number of hyperparameters and its execution time on TC is insignificant, we could have opted for larger search spaces. Nevertheless, to promote fair comparisons with the SoA, we kept the boundaries of the search spaces similar to those of the other methods.

\section{Reproducibility guidelines}

One of the main goals of our research is to guarantee the transparency and replicability of our code and results. We decided to develop our explanation method using the GRETEL\footnote{The original code is on \textcolor{blue}{\url{https://github.com/MarioTheOne/GRETEL}}. Newer versions of the project will be released on \textcolor{blue}{\url{https://github.com/aiim-research/GRETEL}}.} framework. GRETEL \cite{prado2022gretel} is an open-source framework for developing and evaluating Graph Counterfactual Explanation Methods. We provide a compressed folder that contains all the elements that should be included in the GRETEL framework to replicate our results. The folder contains a \texttt{config} sub-folder with the configurations necessary to replicate our experiments and ablation study. Additionally, it contains an \texttt{explainer} sub-folder with the source code of our method and that of the explainer factory, which is necessary to comply with the factory-method design pattern of the framework. The following steps should be followed to use the provided materials to replicate our results.

\begin{enumerate}
    \item Download the source code of the GRETEL framework from \url{https://github.com/MarioTheOne/GRETEL} and place it in a folder with writing permissions.

    \item Copy the content of the \texttt{explainer} folder of the .zip inside the \texttt{src/explainer} folder of the framework and override the existing files.

    \item Copy the content of the \texttt{utils} folder of the .zip inside the \texttt{src/utils} folder of the framework.

    \item To execute a particular experiment, choose the appropriate JSON file inside the config folder of the .zip file and pass it as a parameter to the \texttt{main.py} of the GRETEL framework, along with an integer that represents an identifier that will be assigned to the current run.

    \item Those are all the required steps to execute the experiments. However, verify that all the input/output locations declared in the config JSON files exist.
\end{enumerate}

\section{Extended Qualitative Inspection of Counterfactual Candidates}\label{appendix:qualitative}
\begin{boxA}
\small
\textcolor{black}{The images included in this section are heavy for many viewer applications. We suggest waiting the proper amount of time for their correct visualisation.}
\end{boxA}

\begin{figure*}
    \centering
    \includegraphics[width=\textwidth]{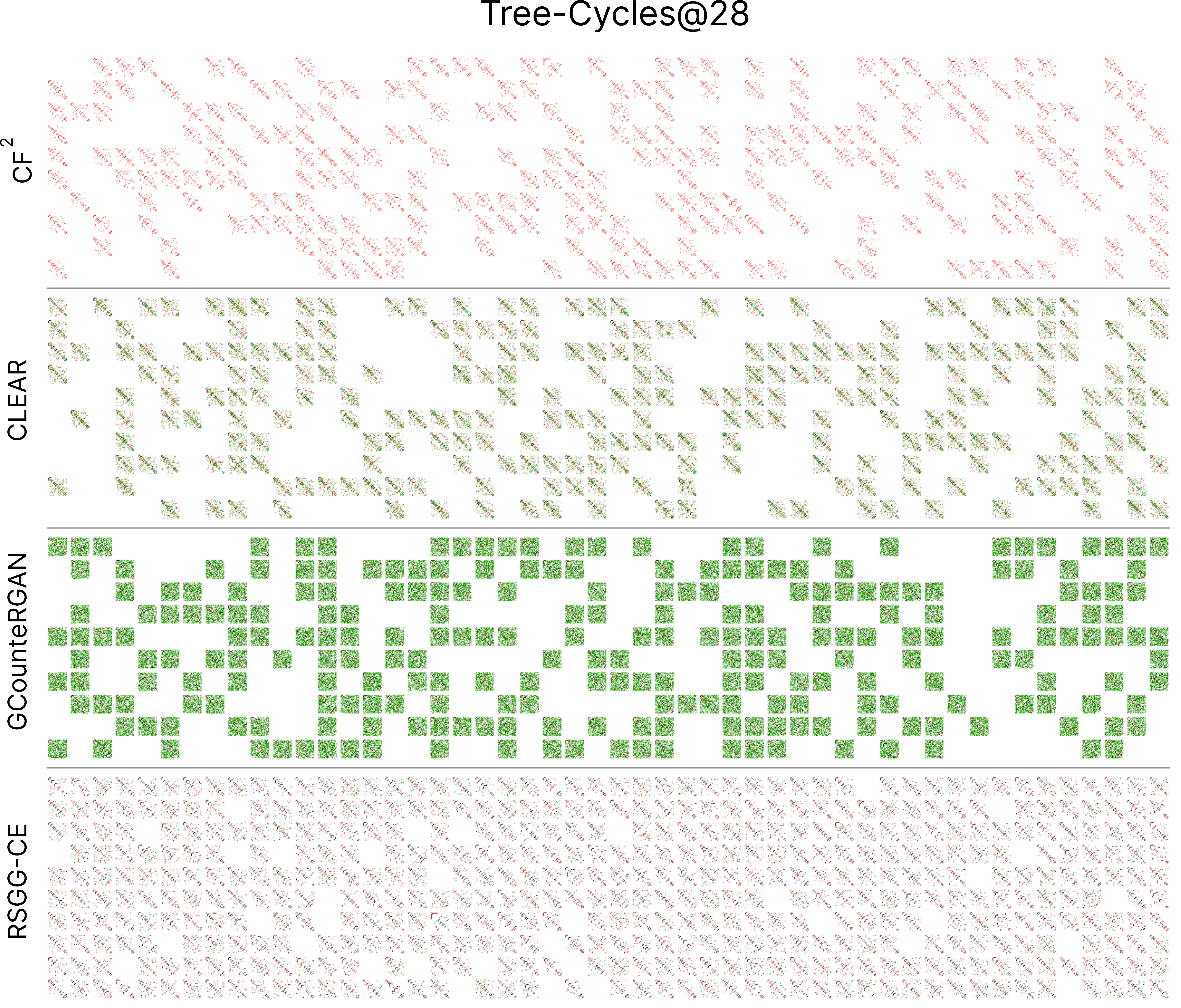}
    \caption{(\textit{best viewed in colour}) Qualitative illustration of the counterfactuals produced by \OUR and SoA approaches on 10 folds (rows) on the test set (columns) in Tree-Cycles with 28 nodes and up to 7 cycles. Each element in the illustration represents the adjacency matrix of the instances. We colour with black those edges that are maintained as in the original instance; with green those that are added; and with red those that are removed. Blank spots are instances where the explainer does not produce a valid counterfactual.}
    \label{fig:pictorial_TC}
\end{figure*}

\begin{figure*}
    \centering
     \includegraphics[width=.85\textwidth]{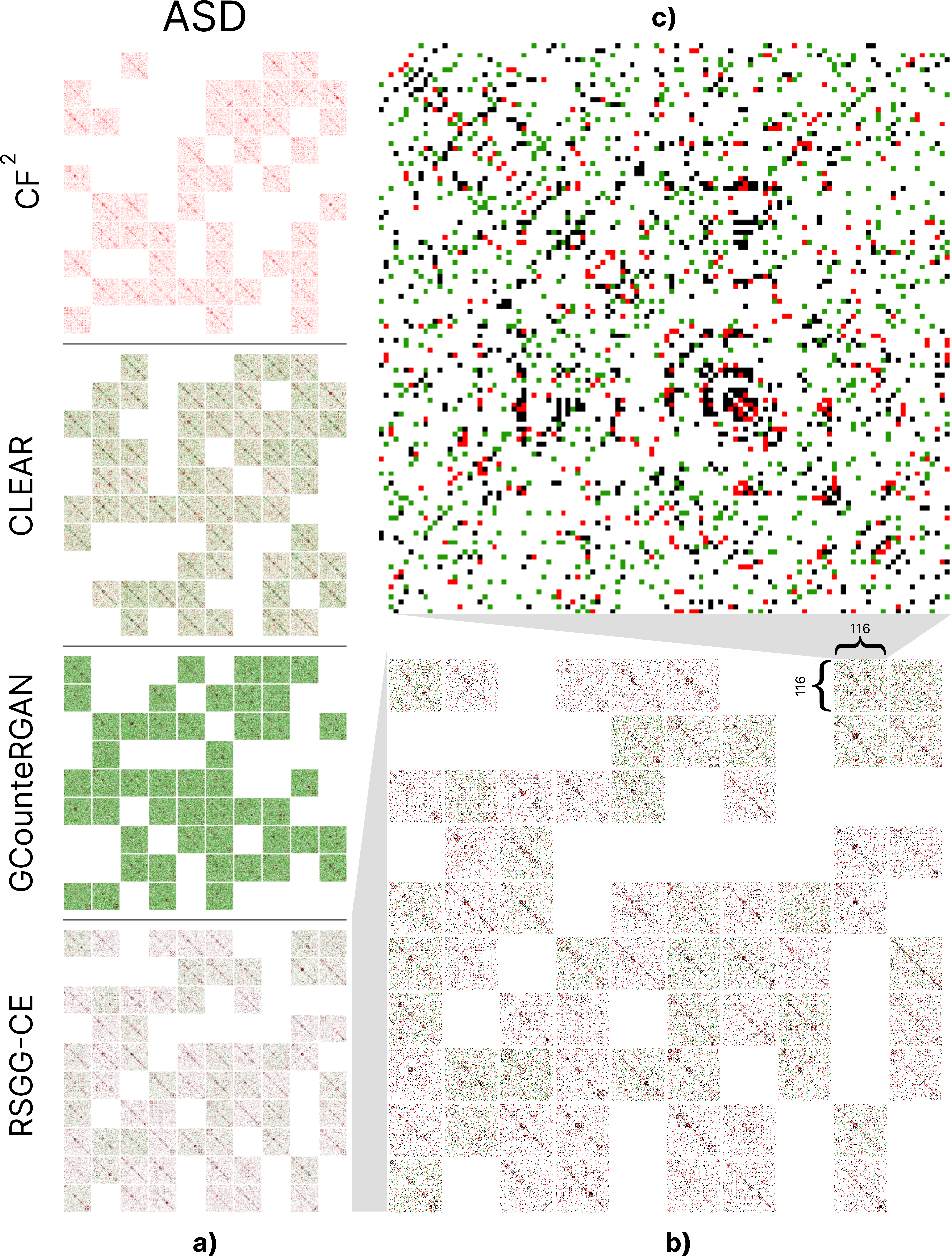}
    \caption{(\textit{best viewed in colour}) Qualitative illustration of the counterfactuals produced by \OUR and SoA approaches on 10 folds (1 for each row) on the test set (columns) in ASD (a). Each element in the illustration represents the adjacency matrix of the instances. We colour with black those edges that are maintained as in the original instance; with green those that are added; and with red those that are removed. Blank spots are instances where the explainer does not produce a valid counterfactual. Zoomed in, we show how \OUR engenders the counterfactual candidates (b). We also illustrate the counterfactual of a specific instance (c).}
    \label{fig:pictorial_ASD}
\end{figure*}

Here, we discuss visually the quality of the counterfactual graphs generated by \OUR and the SoA methods. Figures \ref{fig:pictorial_TC} and \ref{fig:pictorial_ASD} show the generated counterfactuals by  CF$^2$, CLEAR, G-CounteRGAN, and \OUR for the datasets (cf. Sec. \ref{appendix:datasets}) and the comparison conducted in Sec. \ref{sec:experiments}.

\begin{figure*}[t]
\centering
\subfloat[]{%
  \includegraphics[width=.48\linewidth]{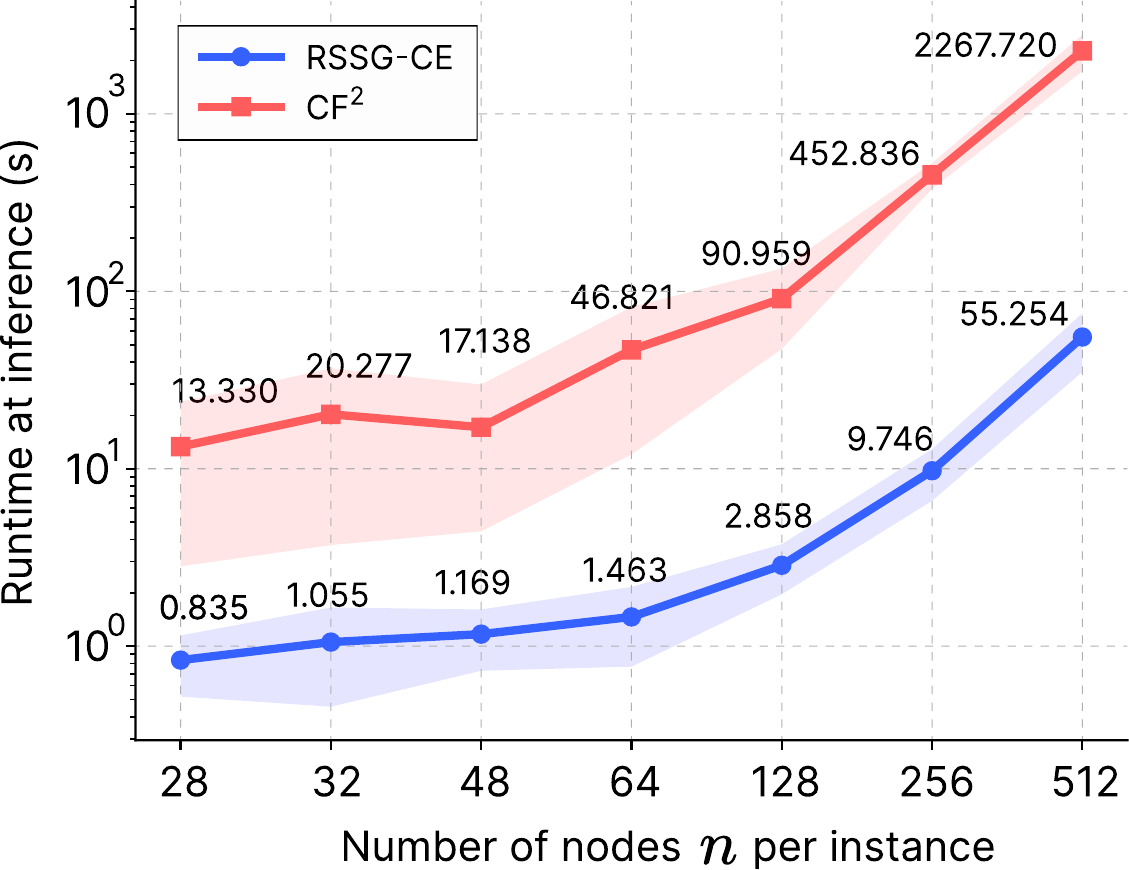}%
  \label{fig:runtime_n_increases}%
}\qquad
\subfloat[]{%
  \includegraphics[width=.45\linewidth]{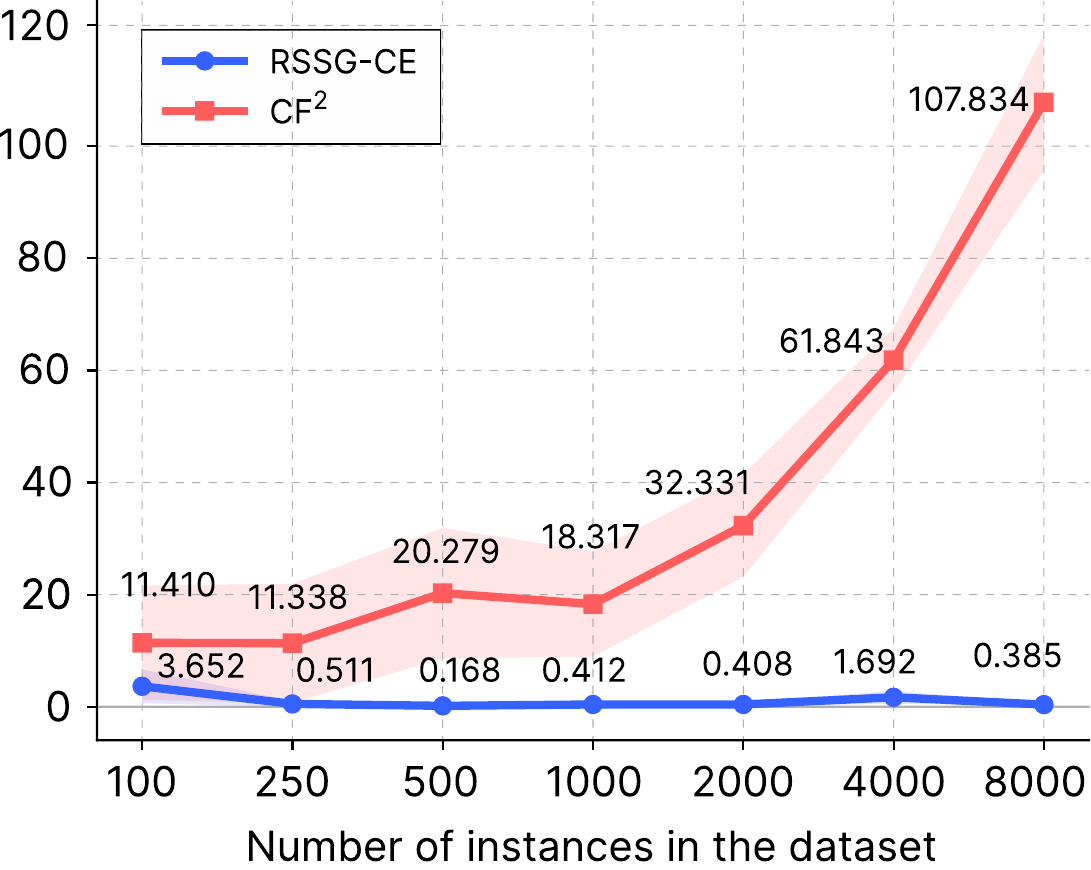}%
  \label{fig:runtime_instances_increase}%
}
\caption{Runtime at inference on TreeCycles when $n$  and number of instances increases, respectively. We use the same configuration of Fig. \ref{fig:ablation_number_of_nodes_per_instance} and \ref{fig:ablation_dataset_size}.}
\end{figure*}

The illustrations are generated directly by processing all the counterfactual candidates engendered for each dataset instance. Each row represents \textit{one fold} where the adjacency matrix of each instance in the test set is converted into a raster image sub-block. As it is possible to notice in Fig. \ref{fig:pictorial_ASD}.c, we have one pixel in the corresponding image sub-block for each edge in the instance.
The pixel is white if the corresponding edge is neither in the input instance nor the counterfactual candidate one. The red pixel denotes an edge in the input instance, but the explainer removes it in the counterfactual candidate. The green pixel denotes an edge not present in the input instance but added in the generated counterfactual candidate. Finally, the black pixel denotes an edge in both the input instance and the counterfactual candidate. 
The white sub-blocks denote that the explainer could not produce a correct counterfactual candidate.
This pictorial provides a quick overview of all the explainers concerning their most preeminent edge operation (i.e., adding or removing edges) and their Correctness. For example, a method with lower Correctness exposes an overall whiter image board since there are more white sub-blocks. 
A green-dominated sub-block indicates that several non-existing edges were generated, while one dominated by red means many removed edges. In general, we expect a good counterfactual candidate to expose a mixture of the three colours, possibly semi-dominated by the black one as it happens for \OUR (see Fig. \ref{fig:pictorial_ASD}.c). CF$^2$ clearly shows its behaviour of removing the factual subgraph in both instances regardless of the class on ASD. However, it is peculiar that the subgraph $\tilde{G}$ corresponds to the original graph $G$, justifying the high GED reported in Table \ref{tab:soa_performance}. CLEAR performs all three types of operations on the original edges.
Although not as naive as CF$^2$'s edge perturbation policy, CLEAR has a higher GED than CF$^2$ and \OUR because it over-generates non-existing edges as shown by its tendency to green in the Figures \ref{fig:pictorial_TC}, \ref{fig:pictorial_ASD}.
G-CounteRGAN, as shown in Table \ref{tab:soa_performance}, is the worst-performing approach, tending to produce a densely connected graph by adding non-existing edges, as evidenced by its tendency to full green in Figs. \ref{fig:pictorial_TC}, \ref{fig:pictorial_ASD}.
In general, \OUR is capable of edge addition/removal operations, but it is guided more by removal ones, as evidenced by its image tendency to be slightly red in Figs. \ref{fig:pictorial_TC}, \ref{fig:pictorial_ASD}. 
Note that because \OUR's produced valid explanation has a low GED (fewer pixels in each sub-block), the partial order sampling strategy engenders a valid counterfactual primarily by examining the original edges.

\begin{figure*}[!h]
\centering
\subfloat[]{%
  \includegraphics[width=.48\linewidth]{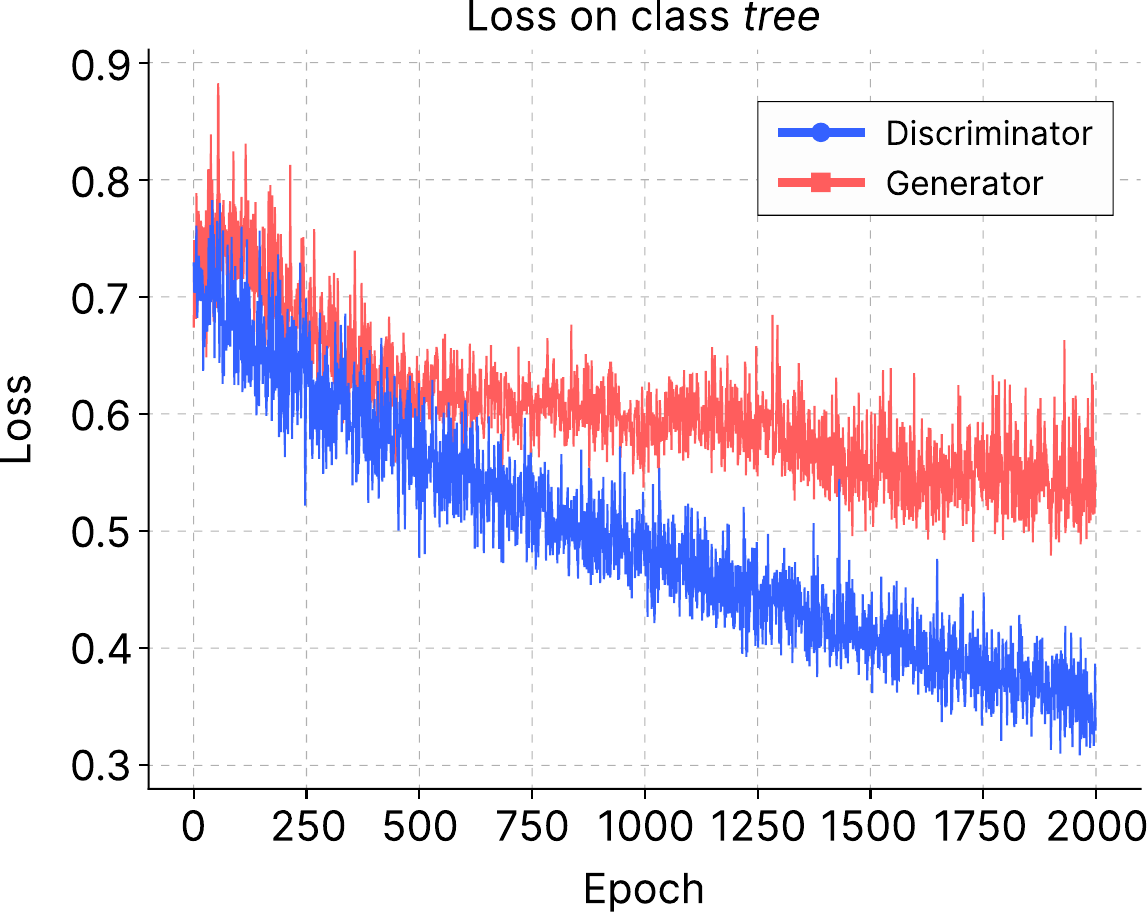}%
  \label{fig:loss_on_cls_tree}%
}\qquad
\subfloat[]{%
  \includegraphics[width=.45\linewidth]{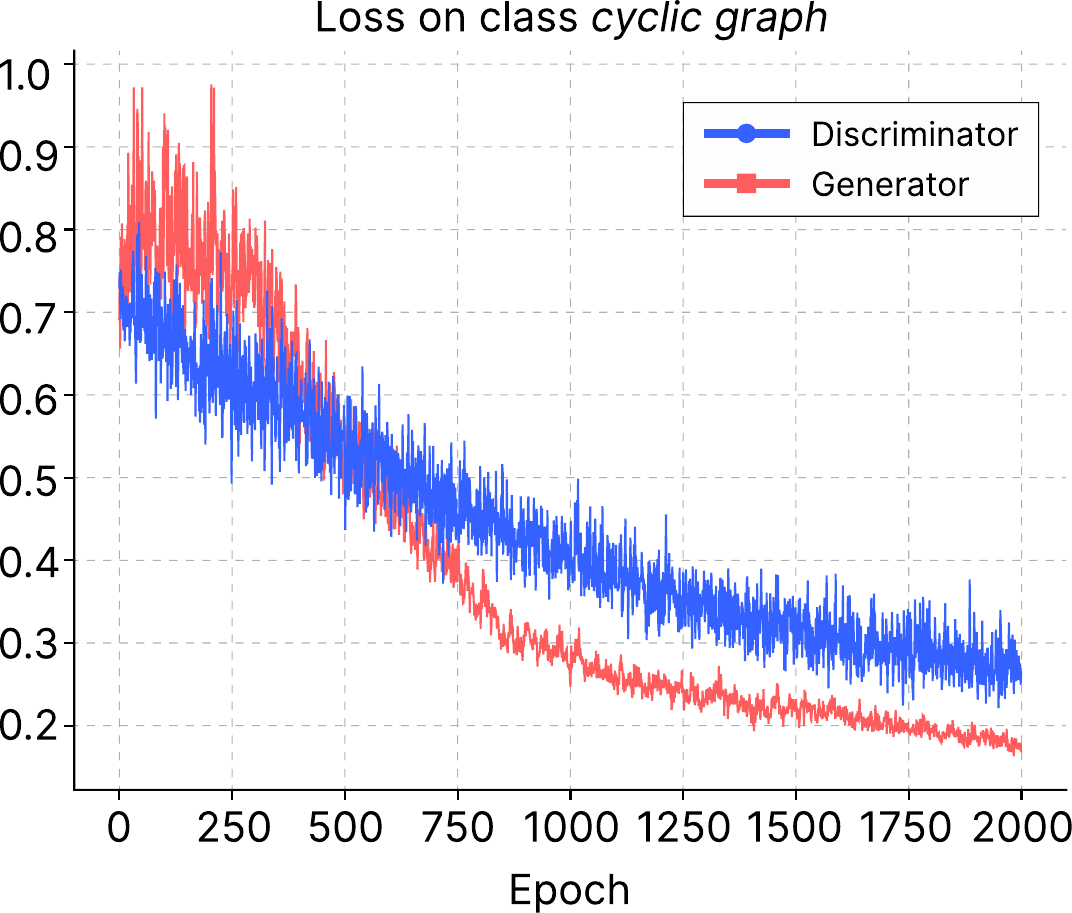}%
  \label{fig:loss_on_cls_graph}%
}
\caption{Loss trend over the epochs for classes \textit{tree} and \textit{cyclic graph}.}
\end{figure*}

\begin{figure*}
    \centering
\includegraphics[width=\textwidth]{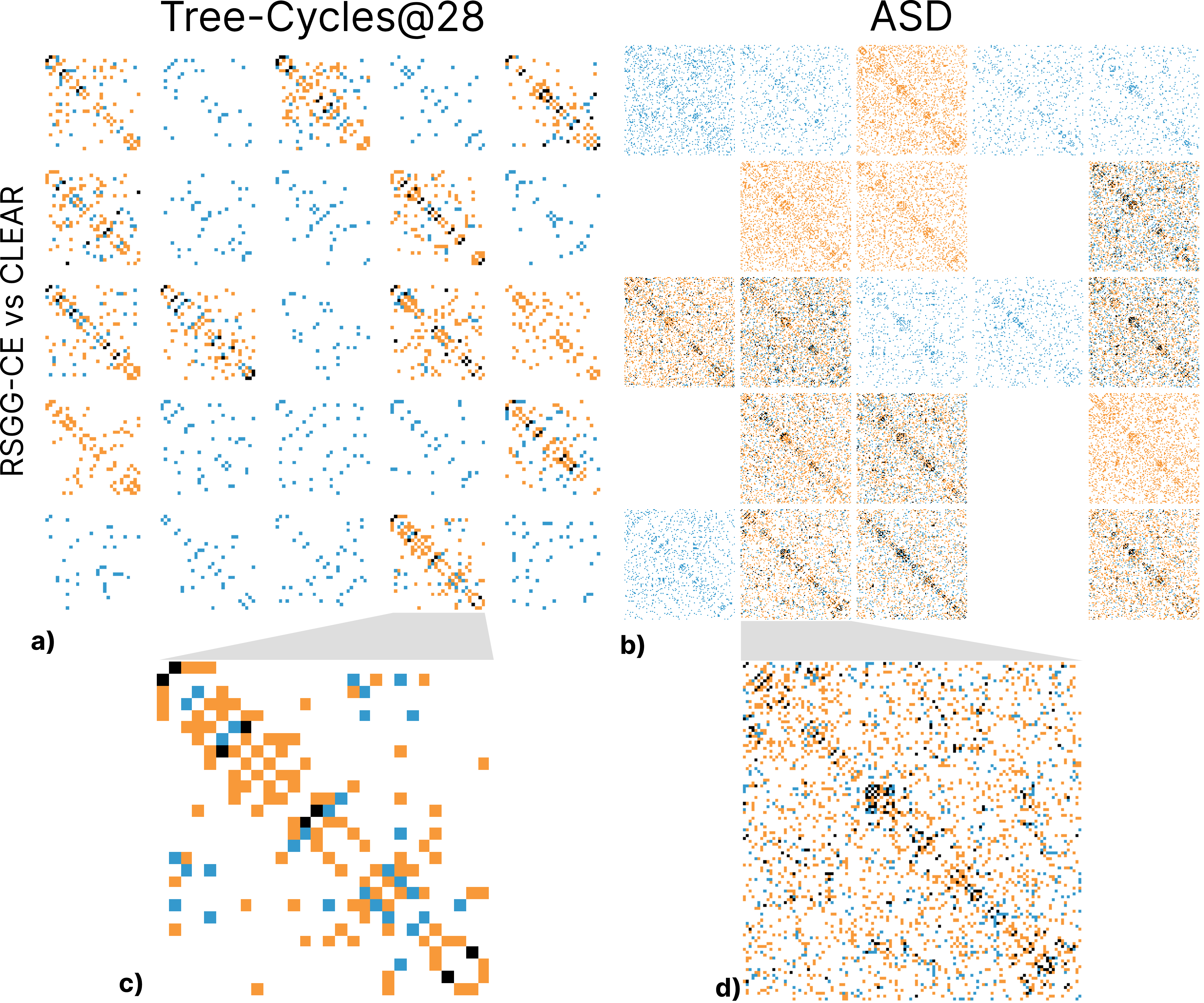}
    \caption{(\textit{best viewed in colour}) Qualitative illustration of the difference between the counterfactual candidates produced by \OUR and CLEAR on five folds (rows) on the randomly chosen five instances of the test set (columns) in Tree-Cycles with 28 nodes and up to 7 cycles (left) and ASD (right). Each element in the illustration represents the adjacency matrix of the instances. We colour with black those edges that are in common between both counterfactuals, with orange edges engendered by CLEAR but not by \OUR, and with blue those edges that \OUR engenders but CLEAR does not. Blank image sub-blocks represent a failure of both explainers to produce valid counterfactuals. Single colour adjacency matrices represent that one of the explainers correctly produces a counterfactual and the other does not: i.e., full orange means that \OUR fails to produce valid counterfactuals, and full blue means that CLEAR fails to produce them.}
    \label{fig:RSSG_vs_CLEAR}
\end{figure*}

Since \OUR and CLEAR  expose the most similar behaviour (very similar colours in Figs. \ref{fig:pictorial_TC}, \ref{fig:pictorial_ASD}), we decided to delve into a direct comparison to have an idea of which is the generation behaviour of one w.r.t. the other. Fig. \ref{fig:RSSG_vs_CLEAR} showcases a comparative view of counterfactual candidates generated by \OUR and CLEAR. For visualisation clarity, we chose here to represent only five randomly selected folds and instances from the previous results. As mentioned above, each image sub-block represents an adjacency matrix of the produced counterfactual candidate. The visual encoding employs colours to highlight differences between the counterfactual outputs of the two explainers. Common edges shared by both counterfactuals are shaded in black, symbolising consensus between the methods.

Additionally, edges exclusively generated by CLEAR and not by \OUR are coloured in orange. On the other hand, edges solely engendered by \OUR, but not CLEAR, are depicted in blue. The illustration also accounts for instances where neither explainer produces a valid counterfactual; these instances remain blank image sub-blocks within the visualisation. Furthermore, instances wherein only one explainer successfully generates a counterfactual are represented by single-colour adjacency matrices. For instance, a matrix entirely orange implies that \OUR fails to produce valid counterfactuals for that instance.

In contrast, a fully blue matrix indicates that CLEAR falls short in generating viable counterfactuals. In detail, one can notice that CLEAR performs a higher amount of edges w.r.t. \OUR (i.e., notice the higher concentration of orange in the adjacency matrices, especially in ASD). Oppositely, \OUR exposes a lower number of edges in total (i.e., the sum of black and blue edges vs the sum of black and orange for CLEAR). This visualisation effectively highlights the differences between \OUR and CLEAR in producing counterfactual candidates across the specified datasets and instances, offering insights into their strengths and weaknesses.

\section{Efficiency and Convergence Analyses}\label{appendix:eff_conv}
We assess the \textit{Runtime} (see Figs. \ref{fig:runtime_n_increases} and \ref{fig:runtime_instances_increase}) at inference time for CF$^2$ and RSGG-CE and \textit{Convergence} (see Figs. \ref{fig:loss_on_cls_tree} and \ref{fig:loss_on_cls_graph}) of RSGG-CE to show that our method can be deployed on large datasets and avoid mode collapse. Notice that RSGG-CE scales perfectly (flat runtime) when the datasets get bigger (\ref{fig:runtime_n_increases}), and it has a linear trend when the graph size increases (\ref{fig:runtime_instances_increase}). As shown in Figs. \ref{fig:loss_on_cls_tree} and \ref{fig:loss_on_cls_graph}, the loss of the discriminator/generator decreases monotonically and stabilises within the same region, avoiding mode collapse that happens, w.l.o.g., if the loss of the discriminator increases whilst that of the generator decreases.

\section{Evaluation Metrics}\label{appendix:eval_metrics}

We adopt the evaluation framework proposed by \cite{prado2022survey}, employing a diverse set of metrics for a comprehensive and fair evaluation. Our evaluation criteria include \textit{Runtime}, \textit{Oracle Calls} \cite{abrate2021counterfactual}, \textit{Correctness} \cite{guidotti2022counterfactual,prado2022gretel}, \textit{Sparsity} \cite{prado2022gretel,yuan2022explainability}, \textit{Fidelity} \cite{yuan2022explainability}, Oracle Accuracy, and \textit{Graph Edit Distance} (GED) \cite{prado2022survey}. Recall that $\Phi: \mathcal{G} \rightarrow Y$ is an oracle where $Y = \{0,1\}$ w.l.o.g.

\noindent\textbf{Runtime} measures the time the explainer takes to generate a counterfactual example. This metric offers an efficient means of evaluating the explainer's performance, encompassing the execution time of the oracle. To ensure fairness, runtime evaluations must be conducted in isolation on the same hardware and software platform.

\noindent\textbf{Oracle Calls} \cite{abrate2021counterfactual} quantifies the number of times the explainer queries the oracle to produce a counterfactual. This metric, akin to runtime, assesses the computational complexity of the explainer, especially in distributed systems. It avoids considering latency and throughput, which are external factors in the measurement.

\noindent\textbf{Oracle Accuracy} evaluates the reliability of the oracle in predicting outcomes. The accuracy of the oracle significantly impacts the quality of explanations, as the explainer aims to elucidate the model's behaviour. Mathematically, for a given input $G$ and true label $y$, accuracy is defined as $\chi(G) = \mathbb{I}[\Phi(G) = y]$.

\noindent\textbf{Correctness} \cite{guidotti2022counterfactual,prado2022gretel} assesses whether the explainer produces a valid counterfactual explanation, indicating a different classification from the original instance. Formally, for the original instance $G$, the counterfactual $G'$, and oracle $\Phi$, correctness is an indicator function $\Omega(G,G') = \mathbb{I}[\Phi(G) \neq \Phi(G')]$.

\noindent\textbf{Sparsity} \cite{yuan2022explainability} gauges the similarity between the input instance and its counterfactual concerning input attributes. If $\mathcal{S}(G,G') \in \mathbb{R}_0^1$ is the similarity between $G$ and $G'$, we adapt the sparsity definition to $\frac{1-\mathcal{S}(G,G')}{|G|}$ for graphs.

\noindent\textbf{Fidelity} \cite{yuan2022explainability} measures the faithfulness of explanations to the oracle, considering correctness. Given the input $G$, true label $y$, and counterfactual $G'$, fidelity is defined as $\Psi(G,G') = \chi(G) - \mathbb{I}[\Phi(G') = y]$. Fidelity values can be $\mathbf{1}$ for the correct explainer and oracle, or $\mathbf{0}$ and $\mathbf{-1}$ indicating issues with either the explainer or the oracle.

\noindent\textbf{Graph Edit Distance (GED)} quantifies the structural distance between the original graph $G$ and its counterfactual $G'$. The distance is evaluated based on a set of actions $\{p_1,p_2,\dots,p_n\} \in \mathcal{P}(G,G')$, representing a path to transform $G$ into $G'$. Each action $p_i$ is associated with a $\omega(p_i)$ cost. GED is computed as 
$$\min_{\{p_1,\dots,p_n\} \in \mathcal{P}(G,G')} \sum_{i = 1}^n \omega(p_i)$$
Preference is given to counterfactuals closer to the original instance $G$, as they provide shorter action paths on $G$ to change the oracle's output. GED offers a global measure and can be complemented by a relative metric like sparsity to assess the explainer's performance across instances.

\section{Towards \OUR\ for Node Classification}\label{sec:node_adapation}

Recall that \OUR\ is devised for graph classification. However, it can be adapted for node classification as follows. The dataset should have instances of form $(v, X, A)$ where $v$ is a target node, and $(X, A)$ is a graph with $v$. Similarly to what we do now, we can divide the dataset according to the classes of the target nodes (e.g., $c = \text{"malicious"}$ and $1-c = \text{"genuine"}$ users in a social network) to train the generator and discriminator appropriately. Now, the generator gets instances $(v_c, X,A)$ and the discriminator those $(v_{1-c}, X,A)$. $\Phi$ must be a node classifier taking in input instances $(v,X,A)$ and predict the label of the node $v$ (e.g., malicious/genuine). The optimisation function in Eq. \ref{eq:res_gan_oracle} should also be modified to reflect the node classification task. We will investigate counterfactual explainability on node classification tasks in the future.

\end{appendix}

\end{document}